\documentclass[journal]{IEEEtran}
\usepackage[utf8]{inputenc} 
\usepackage[T1]{fontenc}    
\usepackage{hyperref}       
\usepackage{url}            
\usepackage{booktabs}       
\usepackage{amsmath, amssymb,amsfonts}       
\usepackage{nicefrac}       
\usepackage{microtype}      
\usepackage{xcolor}         
\usepackage{dsfont}
\usepackage{mathrsfs}
\usepackage[pdftex]{graphicx}
\usepackage{multirow}
\usepackage{cite}
\usepackage{stmaryrd}
\usepackage[ruled,linesnumbered]{algorithm2e}
\usepackage{colortbl}
\usepackage[caption=false]{subfig}
\usepackage{pifont}

\usepackage{bm}

\usepackage{xspace}
\makeatletter
\DeclareRobustCommand\onedot{\futurelet\@let@token\@onedot}
\def\@onedot{\ifx\@let@token.\else.\null\fi\xspace}
\def\eg{\emph{e.g}\onedot} 
\def\ie{\emph{i.e}\onedot}

\def\etal{\emph{et al}\onedot}
\makeatother

\begin{document}
\title{Rethinking Image Forgery Detection via Soft Contrastive Learning and Unsupervised Clustering}

\author{Haiwei~Wu,~Yiming~Chen,~Jiantao~Zhou,~\IEEEmembership{Senior Member,~IEEE},~and~Yuanman~Li,~\IEEEmembership{Senior Member,~IEEE}
\IEEEcompsocitemizethanks{
\IEEEcompsocthanksitem H. Wu is with the School of Computer Science and Engineering, University of Electronic Science and Technology of China, Chengdu 611731, China. Email: haiweiwu@uestc.edu.cn.
\IEEEcompsocthanksitem Y. Chen and J. Zhou are with the State Key Laboratory of Internet of Things for Smart City, and also with the Department of Computer and Information Science, Faculty of Science and Technology, University of Macau, Macau 999078, China. Email: \{yc17486, jtzhou\}@umac.mo. \emph{(Corresponding author: Jiantao Zhou.)}
\IEEEcompsocthanksitem Y. Li is with Guangdong Key Laboratory of Intelligent Information Processing, College of Electronics and Information Engineering, Shenzhen University, Shenzhen 518060, China. Email: yuanmanli@szu.edu.cn.
}}



\maketitle

\begin{abstract}
Image forgery detection aims to detect and locate forged regions in an image. Most existing forgery detection algorithms formulate classification problems to classify pixels into forged or pristine. However, the definition of forged and pristine pixels is only relative within one single image, \eg, a forged region in image A is actually a pristine one in its source image B (splicing forgery). Such a relative definition has been severely overlooked by existing methods, which unnecessarily mix forged (pristine) regions across different images into the same category. To resolve this dilemma, we propose the FOrensic ContrAstive cLustering (FOCAL) method, a novel, simple yet very effective paradigm based on soft contrastive learning and unsupervised clustering for the image forgery detection. Specifically, FOCAL 1) designs a soft contrastive learning (SCL) to supervise the high-level forensic feature extraction in an image-by-image manner, explicitly reflecting the above relative definition; 2) employs an on-the-fly unsupervised clustering algorithm (instead of a trained one) to cluster the learned features into forged/pristine categories, further suppressing the cross-image influence from training data; and 3) allows to further boost the detection performance via simple feature-level concatenation without the need of retraining. Extensive experimental results over six public testing datasets demonstrate that our proposed FOCAL \textbf{\textit{significantly}} outperforms the state-of-the-art competitors by big margins: +24.8\% on \texttt{Coverage}, +18.9\% on \texttt{Columbia}, +17.3\% on \texttt{FF++}, +15.3\% on \texttt{MISD}, +15.0\% on \texttt{CASIA} and +10.5\% on \texttt{NIST} in terms of IoU (see also Fig.~\ref{fig:chart}). The paradigm of FOCAL could bring fresh insights and serve as a novel benchmark for the image forgery detection task. The code is available at \url{https://github.com/HighwayWu/FOCAL}.    
\end{abstract}



\section{Introduction}\label{sec:introduction}

\IEEEPARstart{T}{he} continuous advancement and widespread availability of image editing tools such as Photoshop and Meitu have led to very convenient manipulations of digital images without much domain knowledge. The authenticity of images has thus attracted great attention recently, as maliciously manipulated (forged) images could bring serious negative effects in various fields such as rumor spreading, economic fraud, acquisition of illegal economic benefits, etc.

\begin{figure}[t]
	\begin{center}
		\includegraphics[width=0.99\linewidth]{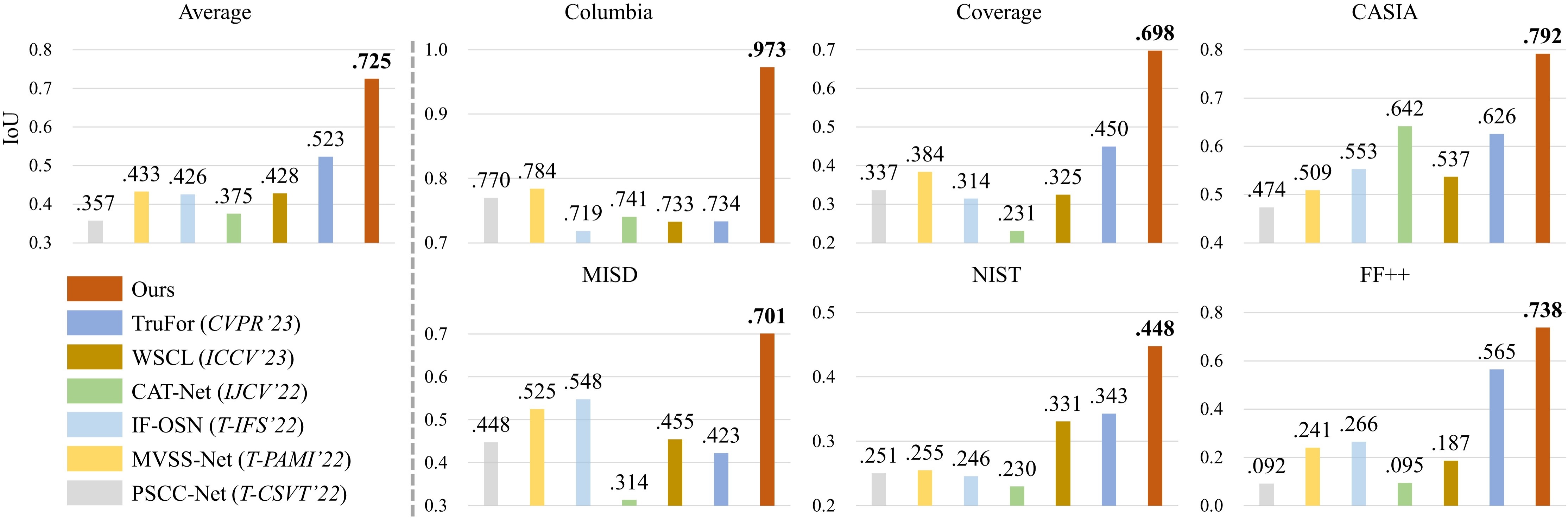}
	\end{center}
	\caption{Our method significantly outperforms several state-of-the-art competing algorithms \cite{mvssnet, catnet, psccnet, ifosn, trufor, wscl} over six \textit{cross-testing} datasets \cite{columbia2006, coverage, casia2013, nist2016, multisp, faceforensics2019}.}
	\label{fig:chart}
\end{figure}

Many forensic methods \cite{noiseprint2020, mvssnet, color2015, trufor, he2012digital, kang2013robust, catnet, psccnet, lyu2014exposing, niloy2023cfl, forensics4_2014, deng2024towards,  ifosn, sun2022dual, bi2023self, yu2024diff, wscl} (and references therein) have been developed to detect and localize forged regions in images, among which the deep learning based schemes offer better performance than the ones relying on hand-crafted features. Several forensic methods are dedicated to detecting specific types of forgery, such as splicing \cite{exif2018}, copy-move \cite{fast2019}, and inpainting \cite{inpaintingFor2021}, while more powerful and practical solutions are for detecting complex and mixed types of forgery, even accompanied with transmission degradation and various post-processing operations \cite{mvssnet, catnet, psccnet, ifosn}. 

\begin{figure}[th!]
	\begin{center}
		\includegraphics[width=0.99\linewidth]{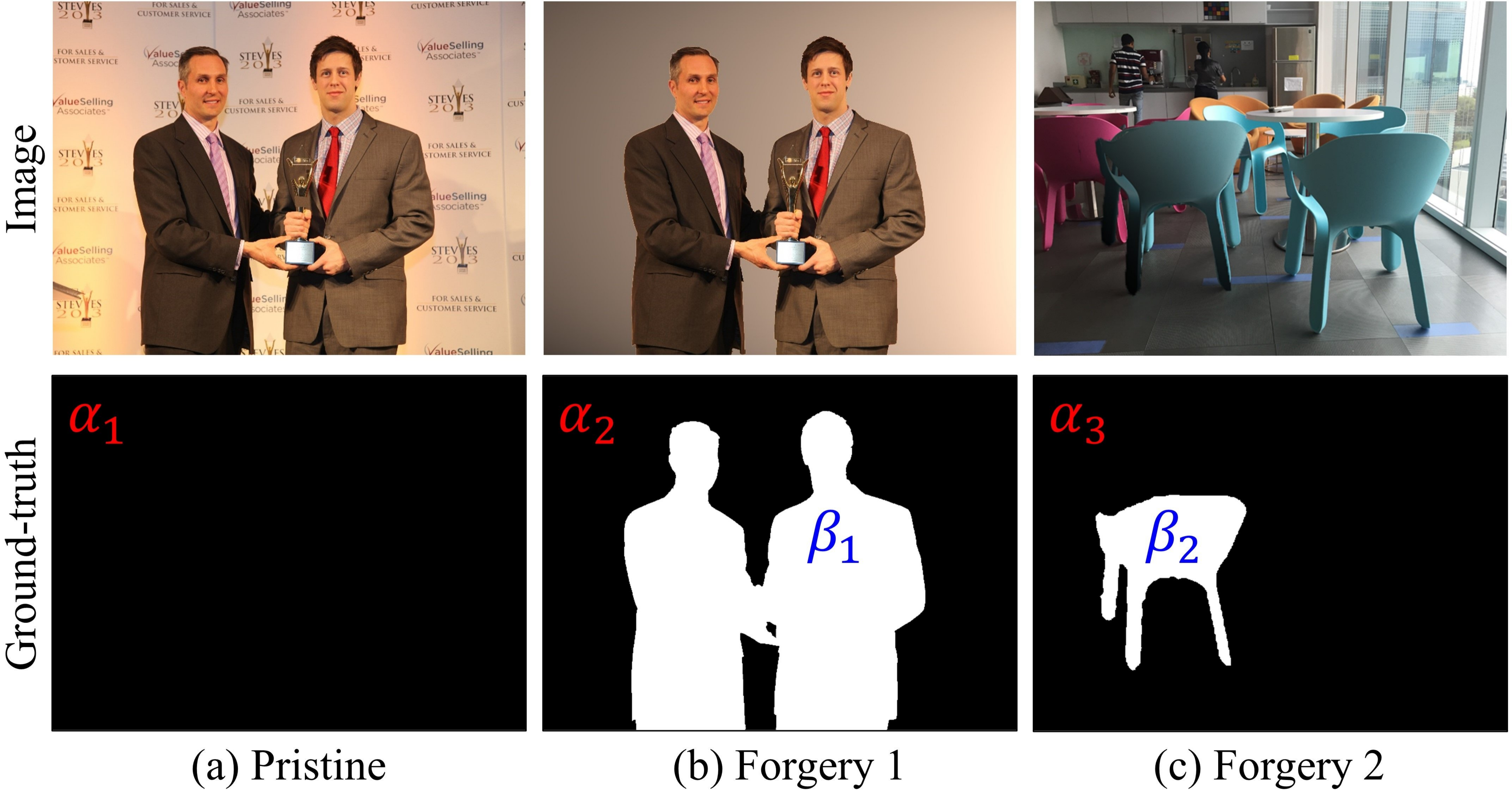}
	\end{center}
	\caption{First row: pristine and forged images. Second row: forgery masks, where pristine ($\alpha_1$, $\alpha_2$ and $\alpha_3$) and forged ($\beta_1$ and $\beta_2$) regions are labeled black and white.}
	\label{fig:head}
\end{figure}

In general, these existing learning-based image forgery detection methods formulate two-class classification problems to classify pixels into forged or pristine. It should be pointed out that the definition of forged and pristine pixels is only \textit{relative} within one single image. For instance, pixels associated with the two persons in Fig.~\ref{fig:head} (a) are pristine, while the same pixels are forged in Fig.~\ref{fig:head} (b), which may lead to \textbf{\textit{label conflict}}. Unfortunately, such a relative definition has been severely overlooked by existing classification-based forgery detection methods, which unnecessarily mix forged (pristine) regions across different images into the same category. In fact, the regions $\alpha_1$, $\alpha_2$, and $\alpha_3$ in Fig.~\ref{fig:head} do not necessarily have similar forensic features, though they belong to the same pristine category (similarly for $\beta_1$ and $\beta_2$). As a result, a classifier could be misled when seeing the same set of pixels are labeled as forged and pristine unfavorably, leading to unstable training and inferior detection performance.

Rethinking the relative definition of forged and pristine pixels inspires us to re-formulate the previously prevailing classification problem, into a new paradigm with contrastive learning and unsupervised clustering. Specifically, we in this work propose the FOrensic ContrAstive cLustering (FOCAL) method, a novel, simple yet effective paradigm for image forgery detection. Firstly, the essence of FOCAL is to directly supervise high-level features by using pixel-level ground-truth forgery mask, explicitly exploiting the above relative definition. Considering that pixel-level ground-truth may cause the so-called \textbf{\textit{label ambiguity}} when supervising the high-level feature learning, we propose a soft contrastive learning (SCL) for FOCAL training. Specifically, SCL introduces optimizable coefficients to meticulously fine-tune the weights of features w.r.t. pristine and forged categories. In addition, another unique characteristic of our designed SCL is the image-by-image supervision, which could effectively avoid the mutual influence (\ie, label conflict) of features across different images in a batch. Further, FOCAL employs an on-the-fly unsupervised clustering algorithm to cluster the learned features into forged/pristine categories, further avoiding the cross-image interference from the training data. Note that here the adopted clustering module does not involve any trainable parameters and hence does not participate in the training process. It is also shown that further performance improvement can be achieved via direct feature-level fusion without the need of retraining.  

Extensive experimental results over six public testing datasets demonstrate that our proposed FOCAL \textit{\textbf{significantly}} outperforms the state-of-the-art competing algorithms \cite{mvssnet, catnet, psccnet, ifosn, trufor, wscl} by big margins: +24.8\% on \texttt{Coverage} \cite{coverage}, +18.9\% on \texttt{Columbia} \cite{columbia2006}, +17.3\% on \texttt{FF++} \cite{faceforensics2019}, +15.3\% on \texttt{MISD} \cite{multisp}, +15.0\% on \texttt{CASIA} \cite{casia2013} and +10.5\% on \texttt{NIST} \cite{nist2016} in terms of IoU. The paradigm of FOCAL could bring fresh insights and serve as a novel benchmark for the image forgery detection task. Our major contributions can be summarized as follows:
\begin{itemize}
	\item We rethink the inherent limitations of classification-based image forgery detection, from the perspective of relative definition of forged/pristine pixels.
	
	\item We design FOCAL, a novel, simple yet effective paradigm based on the proposed SCL and unsupervised clustering for image forgery detection.
	
	\item The proposed FOCAL significantly outperforms several state-of-the-art image forgery detection methods over six (cross-domain) datasets with average gains being 20.2\% in IoU and 10.8\% in F1.

\end{itemize}

The remainder of this paper is organized as follows. Section~\ref{sec:related_work} introduces related works on image forgery detection. Section \ref{sec:focal} gives the details of our proposed FOCAL framework. Experimental results and the analysis are provided in Section~\ref{sec:experiment} and Section~\ref{sec:conclusion} concludes.

\section{Related Works on Image Forgery Detection}\label{sec:related_work}

Classification-based image forgery detection with deep learning has achieved the state-of-the-art performance \cite{mvssnet, catnet, psccnet, ifosn}. CAT-Net \cite{catnet} localizes forged regions through classifying DCT coefficients. PSCC-Net \cite{psccnet} utilizes multi-scale features for forgery detection. Dong \etal \cite{mvssnet} introduced MVSS-Net to jointly extract forged features by multi-view learning. Wu~\etal~\cite{ifosn} designed a robust training framework based on adversarial noise modeling for image forgery detection over online social networks. Recently, Guillaro \etal \cite{trufor} presented TruFor which combines both RGB image and a learned noise-sensitive fingerprint to extract forensic clues. Noticing the limitations of the widely-used cross-entropy loss, some recent works also involve contrastive loss to assist the network training for image forgery detection \cite{noiseprint2020, trufor, niloy2023cfl, wang2022jpeg, yin2022contrastive, zeng2023towards, kong2025pami}.

There are only a few methods trying to detect forgery from the perspective of clustering \cite{bondi2017tampering, lyu2014exposing, trad_clustering2021, trad_clustering2011, trad_clustering2012, trad_clustering2017, trad_clustering2018, trad_clustering2009}, though the performance is much inferior to that of the classification-based ones. This type of method mainly uses a simple clustering algorithm to categorize the image blocks (pixels) into forged and pristine, where various noise features, \eg, image noise level \cite{trad_clustering2017}, camera noise \cite{bondi2017tampering, camera2024wu}, JPEG quantization noise \cite{trad_clustering2021}, were adopted.

Although the aforementioned image forgery detection methods have achieved reasonably good results, their design principles are completely different from our proposed FOCAL in the following aspects: 1) classification-based approaches ignore the relative definition of forged and pristine pixels, and thereby do not take advantage of the unsupervised clustering; 2) those approaches involving clustering almost all work with hand-crafted features, which cannot well represent the forensic traces and are difficult to be generalized to unseen forgery types in cross-domain testing; 3) our framework constructs a feature space that explicitly models nuanced relationships between local regions within each image, allowing the detector to discern subtle inconsistencies without predefined artifacts.

\section{FOCAL for Image Forgery Detection}\label{sec:focal}

\begin{figure*}[th!]
	\centering
	\includegraphics[width=0.8\linewidth]{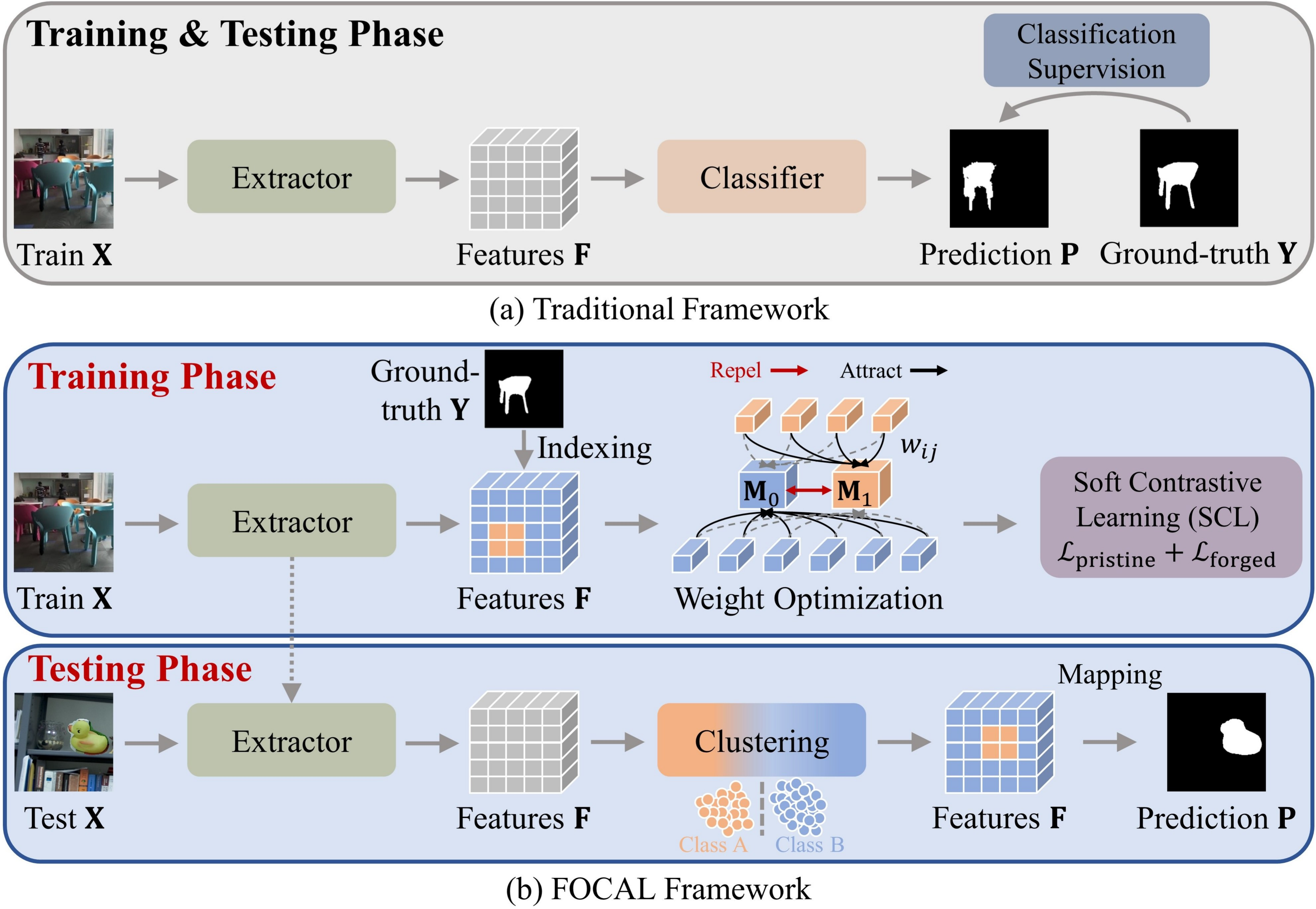}
	\caption{(a) Traditional classification-based forgery detection framework; (b) Our proposed FOCAL framework, which utilizes soft contrastive learning to supervise the training phase, while employing an unsupervised clustering in the testing phase.}
	\label{fig:framework}
\end{figure*}

Before diving into the details of our FOCAL, we introduce the general framework of the traditional classification-based image forgery detection, which consists of two neural networks, namely, \textit{Extractor} and \textit{Classifier}, as shown in Fig.~\ref{fig:framework} (a). Given an input $\mathbf{X} \in \mathbb{R}^{\hat{H} \times \hat{W} \times \hat{C}}$, the extractor first extracts discriminative feature $\mathbf{F} \in \mathbb{R}^{H \times W \times C}$, based on which the classifier generates a predicted binary forgery mask $\mathbf{P} \in \{0,1\}^{H \times W}$. To optimize the network, the cross-entropy loss $\mathcal{L}_{\mathrm{CE}}(\mathbf{P}, \mathbf{Y})$ is usually employed, where $\mathbf{Y} \in \{0, 1\}^{H \times W}$ is the ground-truth forgery mask (1's and 0's for forged and pristine pixels, respectively). Instead of using the classification-based approach, we build a \textit{contrastive-clustering} framework FOCAL (see Fig.~\ref{fig:framework} (b)) for image forgery detection, explicitly exploiting the relative definition of forged/pristine pixels within an image. We now give the details of the FOCAL training via a SCL supervision and FOCAL testing via unsupervised clustering.

\subsection{FOCAL Training via Soft Contrastive Learning (SCL)}

The training procedure of FOCAL is illustrated in the upper part of Fig.~\ref{fig:framework} (b). Upon extracting the high-level features $\mathbf{F}$ from a given input $\mathbf{X}$, we expect features from the same (forged or pristine) regions to attract each other, while features from the different regions repel each other. A natural idea is to use pixel-level ground-truth forgery mask to demarcate the regions (categories) to which these features belong, and then implement a traditional contrastive learning (\eg, NCE~\cite{moco, infonce}) to supervise the update of the feature. However, the potential problem of label ambiguity surfaces at this moment, as some high-level features (\eg, the ones corresponding to the boundary of forged regions) may be extracted from both forged and pristine pixels. To mitigate this label ambiguity, we propose a SCL algorithm, with optimizable weight coefficients to delineate the degree to which a feature pertains to forged or pristine category. As a result, we can construct mean features that can respectively represent the characteristics of forged and pristine regions. Finally, a well-designed soft contrastive supervision is implemented based on the optimized coefficients and mean features.

\begin{figure}[t]
	\centering
	\includegraphics[width=0.99\linewidth]{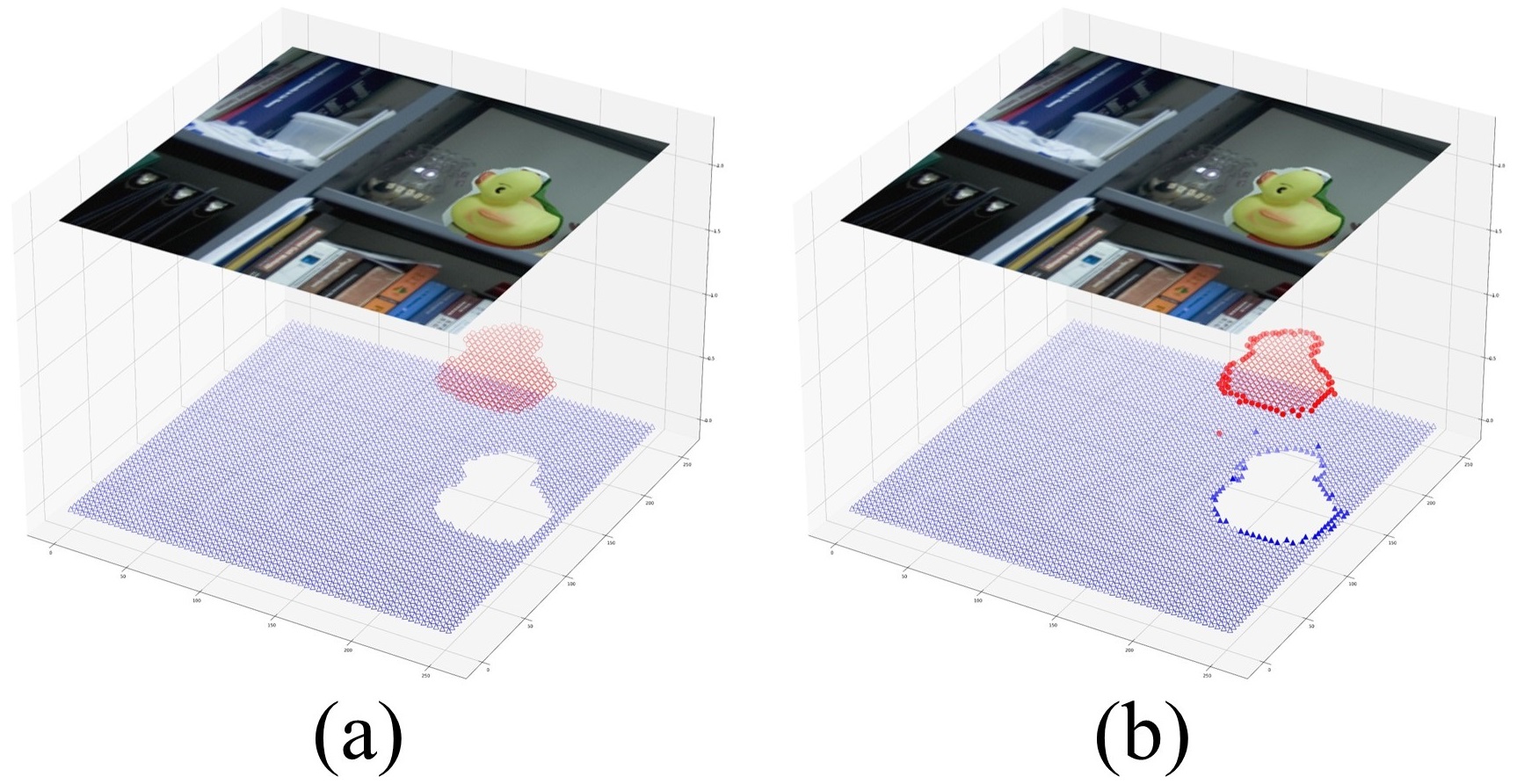}
	\caption{Comparison of initial (a) and optimized (b) $w_{ij}$. The blue and red markers respectively indicate the pristine and forged regions.}
	\label{fig:weight_cmp}
\end{figure}

Specifically, we employ a weight matrix $\left.\mathbf{W}=[w_{ij}]\right.$ wherein each weight coefficient $w_{ij}$ is allocated to feature $\mathbf{F}_i \in \mathbb{R}^{C}$ located at a flattened coordinate $i$ ($i\in\{1,\dots,HW\}$), signifying the degree of $\mathbf{F}_i$ being associated with the $j$-th category ($j \in \{0,1\}$, and $j=0$ for pristine while $j=1$ for forged). According to ground-truth $\mathbf{Y}$, we initialize $w_{i0}=1$ if its index $i$ lies within the pristine regions; conversely, we set $w_{i1}=1$. Evidently, these coefficients are normalized between 0 and 1, and satisfy $\sum_j w_{ij} = 1$. Note that the initialized hard coefficients cannot reflect the soft degree that feature may potentially belong to multiple categories. Letting $\mathbf{M}=[\mathbf{M}_j]$ where $\mathbf{M}_j \in \mathbb{R}^{C}$ represent the feature centers of the $j$-th category (either pristine or forged), we can optimize $w_{ij}$ through minimizing the distance between $\mathbf{F}_i$ and $\mathbf{M}_j$. To this end, we formulate the following constrained optimization problem:
\begin{equation}\label{eq:obj-of-coefficient}
	\begin{split}
		&\underset{\mathbf{W},\mathbf{M}}{\text{minimize}} \ \ J(\mathbf{W},\mathbf{M}), \ \text{where}~ J = \sum_{i,j} w_{ij}^\rho \|\mathbf{F}_i - \mathbf{M}_j\|^2 \\ &\text{subject to}\ \sum_j \mathbf{W}_{\cdot j} =\mathbf{1},
	\end{split}
\end{equation}
where $\|\cdot\|$ is the Euclidean norm, $\mathbf{W}_{\cdot j}$ denotes the $j$-th column vector of $\mathbf{W}$, and $\rho$ is a degree controlling hyper-parameter \cite{fuzzy2012}, empirically set to 2. A commonly adopted optimization approach for constrained problems is Lagrange multiplier method, which begins by constructing the Lagrangian function:
\begin{equation}
	L(\mathbf{W},\mathbf{M},\bm{\lambda}) = \sum_{i,j} w_{ij}^\rho \|\mathbf{F}_i - \mathbf{M}_j\|^2 +  \bm{\lambda} \cdot (\sum_j \mathbf{W}_{\cdot j} -\mathbf{1}),
\end{equation}
where $\bm{\lambda}=[\lambda_i]$ is the vector of Lagrange multipliers. Minimizing $J$ corresponds to identifying the stationary points of $L$, at which the gradients of $L$ w.r.t. $\mathbf{W}$, $\mathbf{M}$ and $\bm{\lambda}$ equal 0. By setting the partial derivative of $L$ w.r.t. $w_{ij}$ to 0, we have:
\begin{equation}\label{eq:partial-wij-1}
	\begin{split}
		& \frac{\partial L}{\partial w_{ij}} = \rho w_{ij}^{\rho-1} \|\mathbf{F}_i - \mathbf{M}_j\|^2 + \lambda_i = 0 \\ \implies & w_{ij} = \Big ( \frac{- \lambda_i}{ \rho \|\mathbf{F}_i - \mathbf{M}_j\|^2} \Big )^{\frac{1}{\rho-1}}.
	\end{split}
\end{equation}
Combining \eqref{eq:partial-wij-1} with the constraints $\frac{\partial L}{\partial \lambda_i} = \sum_j w_{ij} - 1 = 0$, we obtain:
\begin{equation}\label{eq:middle}
	\begin{split}
		& \sum_j \Big ( \frac{- \lambda_i}{ \rho \|\mathbf{F}_i - \mathbf{M}_j\|^2} \Big )^{\frac{1}{\rho-1}} = 1 \\ \implies & (\frac{-\lambda_i}{\rho})^{\frac{1}{\rho-1}} = \Big ( \sum_j \frac{1}{\|\mathbf{F}_i - \mathbf{M}_j\|^2} \Big )^{-\frac{1}{\rho-1}}.
	\end{split}
\end{equation}
Substituting \eqref{eq:middle} into \eqref{eq:partial-wij-1} to eliminate $\lambda_i$, then $w_{ij}$ can be derived as:
\begin{equation}\label{eq:optimized_C}
	w_{ij} = \Big ( \frac{\|\mathbf{F}_i - \mathbf{M}_j\|^2}{\|\mathbf{F}_i - \mathbf{M}_0\|^2} + \frac{\|\mathbf{F}_i - \mathbf{M}_j\|^2}{\|\mathbf{F}_i - \mathbf{M}_1\|^2} \Big )^{-\frac{1}{\rho - 1}}.
\end{equation}
On the other hand, setting the gradient of $L$ w.r.t. $\mathbf{M}_j$ equal to 0 can relieve:
\begin{equation}
	\nabla_{\mathbf{M}_j} L = \sum_j -2 w_{ij}^\rho (\mathbf{F}_i - \mathbf{M}_j) = 0,
\end{equation}
then $\mathbf{M}_j$ can be written as:
\begin{equation}\label{eq:optimized_M}
	\mathbf{M}_j = \frac{\sum_i w_{ij}^\rho \mathbf{F}_i}{\sum_i w_{ij}^\rho}.
\end{equation}
Since the variables of \eqref{eq:optimized_C} and \eqref{eq:optimized_M} are intertwined with each other, we adopt the alternating iteration method to continuously update $w_{ij}$ and $\mathbf{M}_j$ until the objective function $J$ converges. An intuitive exposition of $w_{ij}$ is presented in Fig.~\ref{fig:weight_cmp}, where (a) represents the initial $w_{ij}$ and (b) corresponds to the optimized one. It is evident that $w_{ij}$ embodies the label ambiguity of features situated at the boundary between pristine and forged categories.

\begin{figure}[t]
	\centering
	\includegraphics[width=0.73\linewidth]{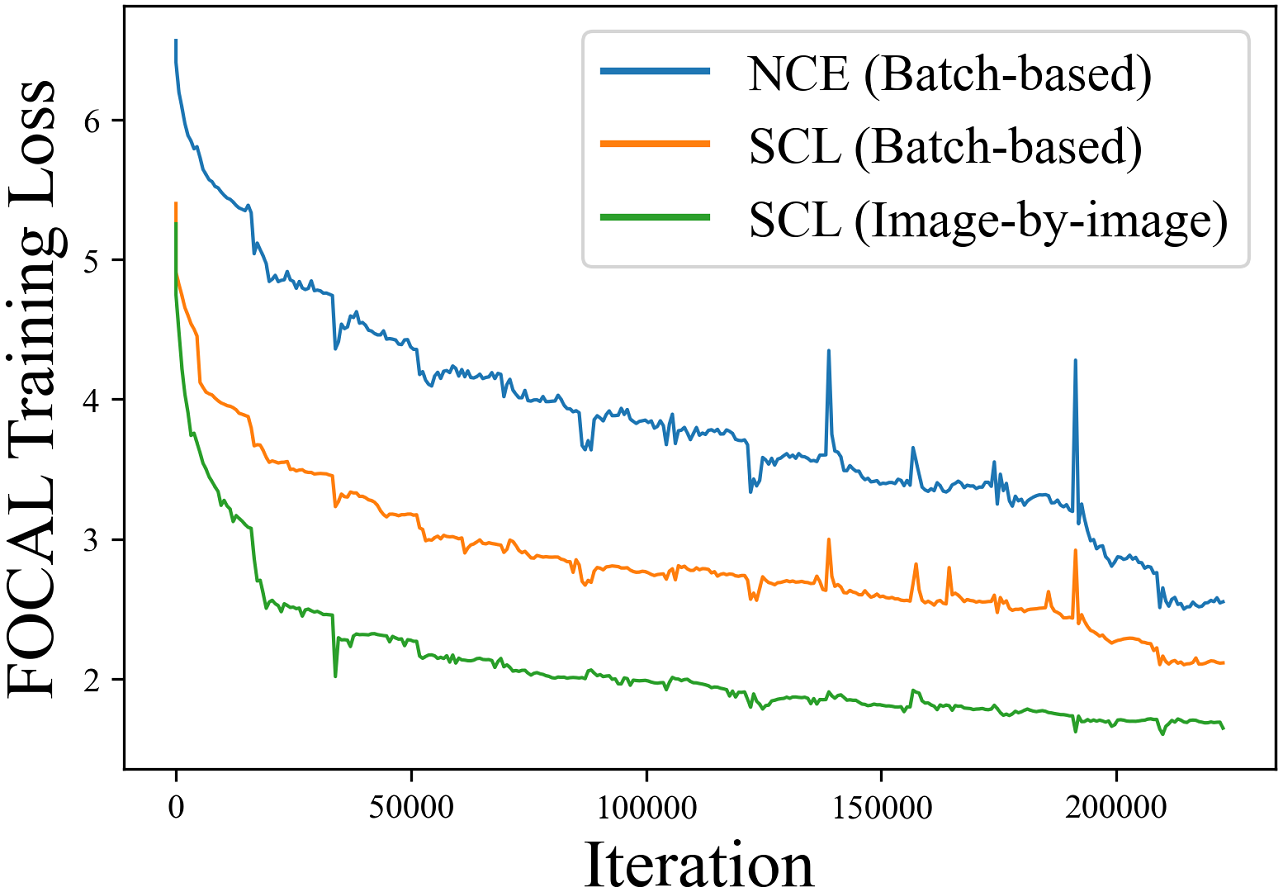}
	\caption{Training loss curves of the NCE baseline (blue), our proposed $\mathcal{L}_{\mathrm{SCL}}$ in batch-based (orange) and image-by-image (green), respectively.}
	\label{fig:loss_cmp}
\end{figure}

After mitigating the issue of label ambiguity, we propose the SCL based on an improved NCE loss to implement the soft contrastive supervision in FOCAL, with the aim of augmenting intra-class similarity while diminishing inter-class similarity. By designating $\mathbf{M}_0$ as the query, $\mathcal{L}_{\mathrm{pristine}}$ can be defined as:
\begin{equation}\label{eq:improved_infoNCE}
	\mathcal{L}_{\mathrm{pristine}} = -\log \frac{\frac{1}{HW} \sum_i \exp(\mathbf{M}_0 \cdot w_{i0}\mathbf{F}_i / \tau)}{\exp(\mathbf{M}_0 \cdot \mathbf{M}_1 / \tau)},
\end{equation}
where $\tau$ is a temperature hyper-parameter \cite{wu2018unsupervised}. In \eqref{eq:improved_infoNCE}, the numerator quantifies the similarity between pristine features, while the denominator characterizes the similarity between pristine and forged central features. Similarly, we can obtain $\mathcal{L}_{\mathrm{forged}}$ from the forged feature perspective by setting the query as $\mathbf{M}_1$:
\begin{equation}
	\mathcal{L}_{\mathrm{forged}} = -\log \frac{\frac{1}{HW} \sum_i \exp(\mathbf{M}_1 \cdot w_{i1}\mathbf{F}_i / \tau)}{\exp(\mathbf{M}_1 \cdot \mathbf{M}_0 / \tau)}.
\end{equation}
The overall SCL training loss then becomes $\mathcal{L}_{\mathrm{SCL}} = \mathcal{L}_{\mathrm{pristine}} + \mathcal{L}_{\mathrm{forged}}$. Compared with the traditional NCE loss, our improved $\mathcal{L}_{\mathrm{SCL}}$ involves all the positive keys in each loss calculation by taking the expectation of the dot product of $\mathbf{M}_j$ with the weighted features $w_{ij}\mathbf{F}_i$'s. This would facilitate the optimization process, as also be verified by the loss curves in Fig.~\ref{fig:loss_cmp}. As will be clear in Sec.~\ref{sec:experiment}, this new loss $\mathcal{L}_{\mathrm{SCL}}$ leads to significant performance gains over the traditional NCE loss.

It should be emphasized that the supervision in the training phase is implicitly conducted between the ground-truth forgery mask $\mathbf{Y}$ and the extracted feature $\mathbf{F}$, while no predicted forgery mask is generated. Furthermore, for each image in the forward mini-batch, $\mathcal{L}_{\mathrm{SCL}}$ is calculated in an image-by-image manner (one-by-one), rather than over the entire batch, and is then summed up to calculate the overall loss. To be more concrete, given a mini-batch features $\{\mathbf{F}^{(1)}, \mathbf{F}^{(2)}, \cdots, \mathbf{F}^{(B)}\}$, the overall soft contrastive loss in this mini-batch $\mathcal{L}_{\mathrm{SCL-IBI}}$ is:
\begin{equation}\label{eq:overallLoss}
	\mathcal{L}_{\mathrm{SCL-IBI}} = \frac{1}{B} \sum_{b=1}^B \mathcal{L}_{\mathrm{SCL}}(\mathbf{F}^{(b)}).
\end{equation} Note that in \eqref{eq:overallLoss}, the mini-batch features are \textit{not} merged to compute an overall $\mathcal{L}_{\mathrm{SCL}}$, avoiding the cross-image influence from the training data. This total loss designed under the guidance of relative definition of forged/pristine pixels is vastly different from those in \cite{simclr, trufor, moco, niloy2023cfl}, where loss computation is conducted at the batch-level. To further justify the rationality of \eqref{eq:overallLoss}, we plot the contrastive loss curves of the traditional batch-based and our image-by-image one in Fig.~\ref{fig:loss_cmp}. It can be clearly seen that the image-by-image design of the loss function (green line) not only leads to much faster convergence, but also makes the optimization much more stable. Particularly, the high-amplitude impulses detected in the blue and orange lines indicate that there might be serious conflicts in the associated batch of images, \eg, a situation similar to the case of Fig.~\ref{fig:head} (a) and (b), where conflicting labels are presented.

Eventually, the well-trained extractor will be used in the FOCAL testing phase. As expected and will be verified experimentally, our proposed SCL loss with image-by-image supervision significantly improves the image forgery detection performance.

\subsection{FOCAL Testing via Unsupervised Clustering}

We now are ready to present the details on the FOCAL testing phase. The crucial issue is how to map the extracted features into a predicted forgery mask. Compared to traditional frameworks using trained classifiers (see Fig.~\ref{fig:framework} (a)), we propose to employ an unsupervised online-learning algorithm (see the bottom half of Fig.~\ref{fig:framework} (b)). As aforementioned, the definition of forged and pristine pixels is only relative within one single image, and can be hardly generalized across different images. This explains why the previous classification-based approaches do not offer satisfactory detection results, as the classifier trained from training data may not be able to infer the unseen testing data.   

Therefore, it would be a wiser solution to map the features of different images to the final forgery mask \textit{separately}. To this end, we adopt an on-the-fly clustering algorithm. Specifically, we employ HDBSCAN \cite{ester1996density} with minimum cluster size empirically set as 200 to cluster $\mathbf{F}$, and label the cluster with the most elements as pristine (otherwise forged), implicitly assuming that forged pixels only occupy a relatively smaller portion. Features $\mathbf{F}$ extracted by our SCL loss with image-by-image supervision may be already very discriminative, making an unsupervised algorithm sufficient to handle the clustering task. The performance comparison using different clustering algorithms is deferred to experimental results.

\textit{Remark:} One feasible approach involves integrating trainable clustering (e.g., differentiable K-means \cite{fard2020deep}) with contrastive loss to form an end-to-end paradigm, enabling joint optimization of intermediate features $\mathbf{F}$ and clustering results. However, we experimentally found that this approach not only incurs significantly increased training time but also fails to yield observable performance improvements. Despite this, exploring hybrid frameworks that balance end-to-end discriminability with clustering robustness remains an open and valuable direction for future research. In summary, end-to-end training paradigm is abandoned due to its unnoticeable performance improvement and additional memory overhead.

\begin{figure}[t!]
	\begin{center}
		\includegraphics[width=0.98\linewidth]{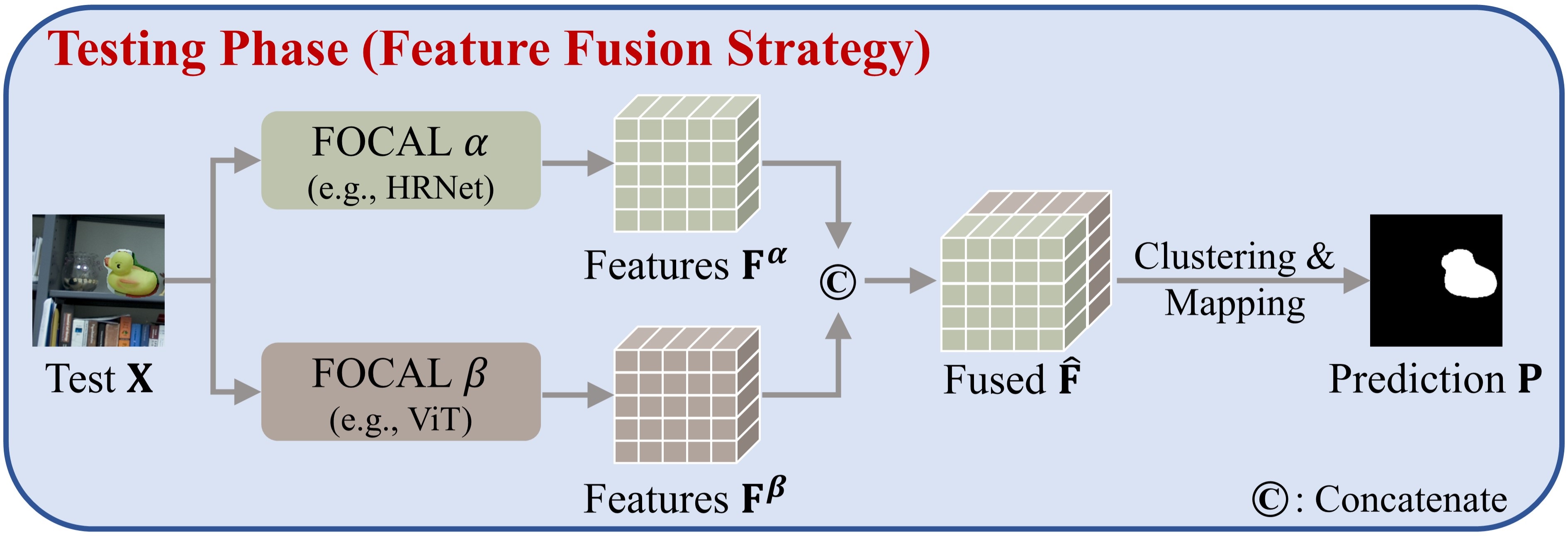}
	\end{center}
	\caption{Feature-level fusion for boosting the detection performance of FOCAL. Retraining is not needed.}
	\label{fig:fusion_framework}
\end{figure}

\begin{table}[t!]
	\caption{Dataset Statistics. SP, CM, SW, PP are short for splicing, copy-move, software, and post-processing, respectively.}
	\centering
	\scalebox{0.8}{
		\begin{tabular}{l|ccccccc}
			\hline
			\hline
			\multirow{2}{*}{Datasets} & \multirow{2}{*}{\#Data} & \multicolumn{5}{c}{Forgery Types} & Resolution \\
			\cline{3-7} & & SP & CM & SW & PP & GAN & (Average) \\
			\cline{1-8}
			Training Datasets \\
			-\texttt{SP-COCO} \cite{catnet} & 200K & \checkmark & & & \checkmark &  & $640\times480$ \\
			-\texttt{CM-COCO} \cite{catnet} & 200K &  & \checkmark & & \checkmark &  & $640\times640$ \\
			-\texttt{CM-RAISE} \cite{catnet} & 200K &  & \checkmark & & \checkmark &  & $512\times512$ \\
			-\texttt{CM-C-RAISE} \cite{catnet} & 200K &  & \checkmark & & \checkmark &  & $512\times512$ \\
			-\texttt{CASIA-v2} \cite{casia2019} & 5105 & \checkmark & \checkmark & \checkmark & \checkmark &  & $384\times256$\\
			-\texttt{IMD2020} \cite{imd2020} & 2010 & \checkmark & \checkmark & \checkmark & \checkmark &  & $1920\times1200$\\
			\hline
			Testing Datasets \\
			-\texttt{Coverage} \cite{coverage} & 100 & & \checkmark & \checkmark & \checkmark &  & $520\times430$ \\
			-\texttt{Columbia} \cite{columbia2006} & 160 & \checkmark & & & &  & $1152\times768$ \\
			-\texttt{NIST} \cite{nist2016} & 540 & \checkmark & \checkmark & \checkmark & \checkmark &  & $5616\times3744$\\
			-\texttt{CASIA} \cite{casia2013} & 920 & \checkmark & \checkmark & \checkmark & \checkmark &  & $384\times256$ \\
			-\texttt{MISD} \cite{multisp} & 227 & \checkmark & & & \checkmark &  & $384\times256$ \\
			-\texttt{FF++} \cite{faceforensics2019} & 1000 & & & & \checkmark & \checkmark & $480\times480$ \\
			\hline
			\hline
		\end{tabular}
	}
	\label{tab:datasets}
\end{table}

\subsection{Feature Fusion Strategy}
We now show that the performance of the standalone FOCAL can be further improved through simple yet effective feature fusion strategy. Fig.~\ref{fig:fusion_framework} gives an example of fusing two FOCAL $\alpha$ and FOCAL $\beta$ with distinct backbones (\eg, HRNet \cite{hrnet} or ViT \cite{vit}). The fused feature can be readily obtained from direct concatenation, namely,         
\begin{equation}\label{eq:fusion}
	\hat{\mathbf{F}} = \mathrm{Concat}(\mathbf{F}^\alpha, \mathbf{F}^\beta),
\end{equation} where $\mathbf{F}^\alpha$ and $\mathbf{F}^\beta$ are extracted features by FOCAL $\alpha$ and FOCAL $\beta$, respectively, and need to be scaled to the same resolutions. Prediction results can then be generated by the subsequent clustering and mapping accordingly. As will be validated through experiments, the above feature-level fusion significantly outperforms the naive result-level fusion \cite{ensemble}. Also, such a feature fusion strategy can be easily extended to cases with more than two FOCAL networks, and there is no retraining involved.

\begin{table*}[t!]
	\centering
	\caption{Quantitative comparison of detection results using F1 and IoU as criteria. $^\dagger$: retrained versions with CAT-Net datasets. The best results are in \textbf{bold} and the second best results (excluding FOCAL variants) are \underline{underlined}.}
	\scalebox{1.15}{
		\begin{tabular}{l|cc|cc|cc|cc|cc|cc|cc}
			\hline
			\hline
			\multirow{2}{*}{Methods} & \multicolumn{2}{c|}{\texttt{Columbia}} & \multicolumn{2}{c|}{\texttt{Coverage}} & \multicolumn{2}{c|}{\texttt{CASIA}} & \multicolumn{2}{c|}{\texttt{MISD}} & \multicolumn{2}{c|}{\texttt{NIST}} & \multicolumn{2}{c|}{\texttt{FF++}} & \multicolumn{2}{c}{Mean} \\
			\cline{2-15} & F1 & IoU & F1 & IoU & F1 & IoU & F1 & IoU & F1 & IoU & F1 & IoU & F1 & IoU\\
			\cline{1-15}
			Lyu-NOI \cite{lyu2014exposing} & .522 & .150 & .481 & .125 & .356 & .095 & .507 & .199 & .478 & .026 & .496 & .071 & .473 & .111 \\
			PCA-NOI \cite{trad_clustering2017} & .539 & .168 & .529 & .125 & .472 & .093 & .517 & .150 & .460 & .046 & .523 & .108 & .507 & .115 \\
			PSCC-Net \cite{psccnet} & .577 & .480 & .655 & .337 & .716 & .409 & .746 & .448 & .300 & .078 & .509 & .092 & .584 & .307 \\
			PSCC-Net$^\dagger$ \cite{psccnet} & .850 & .770 & .584 & .179 & .753 & .474 & .735 & .403 & .632 & .251 & .518 & .068 & .679 & .357 \\
			MVSS-Net \cite{mvssnet} & .766 & .591 & .700 & .384 & .707 & .396 & .803 & .525 & .621 & .243 & .553 & .127 & .691 & .378 \\
			MVSS-Net$^\dagger$ \cite{mvssnet} & \underline{.888} & \underline{.784} & .690 & .356 & .770 & .509 & .765 & .450 & .635 & .255 & .633 & .241 & .730 & .433 \\
			IF-OSN \cite{ifosn} & .766 & .612 & .561 & .178 & .741 & .465 & \underline{.811} & \underline{.548} & .639 & .246 & .628 & .266 & .691 & .386 \\
			IF-OSN$^\dagger$ \cite{ifosn} & .846 & .719 & .651 & .314 & .828 & .553 & .765 & .521 & .608 & .226 & .607 & .222 & .717 & .426 \\
			WSCL \cite{wscl} & .726 & .595 & .538 & .157 & .748 & .523 & .739 & .422 & .591 & .334 & .532 & .152 & .646 & .364 \\
			WSCL$^\dagger$ \cite{wscl} & .825 & .733 & .669 & .325 & .802 & .537 & .781 & .455 & .635 & .331 & .671 & .187 & .731 & .428 \\
			CAT-Net \cite{catnet} & .864 & .741 & .614 & .231 & \underline{.846} & \underline{.642} & .665 & .314 & .620 & .230 & .534 & .095 & .690 & .375 \\
			TruFor \cite{trufor} & .821 & .734 & \underline{.741} & \underline{.450} & .835 & .626 & .746 & .423 & \underline{.688} & \underline{.343} & \underline{.817} & \underline{.565} & \underline{.774} & \underline{.523} \\
			\hline
			FOCAL (HRNet) & .958 & .925 & .762 & .521 & .866 & .708 & .851 & .638 & .708 & .402 & .835 & .603 & .830 & .633  \\
			FOCAL (ViT) & .983 & .975 & .837 & .650 & .889 & .758 & .875 & .662 & .725 & .433 & .848 & .632 & .860 & .685  \\
			FOCAL (Fusion) & \textbf{.985} & \textbf{.973} & \textbf{.866} & \textbf{.698} & \textbf{.904} & \textbf{.792} & \textbf{.892} & \textbf{.701} & \textbf{.741} & \textbf{.448} & \textbf{.902} & \textbf{.738} & \textbf{.882} & \textbf{.725} \\
			\hline
			\hline
		\end{tabular}
	}
	\label{tab:cmp}
\end{table*}

\section{Experimental Results}\label{sec:experiment}
In this section, we first present the detailed experimental settings. Then, image forgery detection/localization results on six public testing datasets are reported and compared with those of several state-of-the-art algorithms. Finally, extensive ablation studies and further analysis are conducted.

\subsection{Settings}
\subsubsection{Training Datasets} We train the FOCAL using the \textit{same} training dataset as \cite{catnet, trufor}. This training dataset contains over 800K forged images from \texttt{SP-COCO}~\cite{catnet}, \texttt{CM-COCO}~\cite{catnet}, \texttt{CM-RAISE}~\cite{catnet}, \texttt{CM-C-RAISE}~\cite{catnet}, \texttt{CASIA-v2}~\cite{casia2019}, and \texttt{IMD2020}~\cite{imd2020}. Specifically, \texttt{CASIA-v2} is a widely-adopted dataset that contains various \textit{multi-source} splicing and copy-move forgeries, while \texttt{IMD2020} collects real-world manipulated images from the Internet. Considering the insufficient numbers of images in these two datasets, Kwon \etal \cite{catnet} utilized splicing and copy-move methods to produce a large amount of forged images based on pristine datasets \texttt{COCO} \cite{coco2014} and \texttt{RAISE} \cite{raise}. To better mimic the distribution of real-world images, a variety of post-processing operations such as resizing, rotation, and compression are involved.

\subsubsection{Testing Datasets} Following \cite{mvssnet, catnet, ifosn, trufor, wscl}, six commonly-used datasets are adopted for testing, namely, \texttt{Columbia} \cite{columbia2006}, \texttt{Coverage} \cite{coverage}, \texttt{CASIA} \cite{casia2013}, \texttt{NIST} \cite{nist2016}, \texttt{MISD} \cite{multisp}, and \texttt{FF++} \cite{faceforensics2019}. These testing datasets encompass a plethora of highly sophisticated forgeries, \eg, \texttt{MISD} comprising multi-source forgeries, and \texttt{FF++} harboring faces synthesized via GANs \cite{goodfellow2014generative}. Note that \textbf{NO} overlap exists between the training and testing datasets, aiming to simulate the practical situation and evaluate the generalization of the forgery detection algorithms. Full statistics of the involved datasets are detailed in Table~\ref{tab:datasets}.

\subsubsection{Competitors} The following state-of-the-art learning-based image forgery detection algorithms PSCC-Net \cite{psccnet}, MVSS-Net \cite{mvssnet}, IF-OSN \cite{ifosn}, WSCL \cite{wscl}, CAT-Net \cite{catnet}, and TruFor \cite{trufor} are selected as comparative methods. Their released codes can be found in their official links \footnote{https://github.com/proteus1991/PSCC-Net} \footnote{ https://github.com/dong03/MVSS-Net} \footnote{ https://github.com/HighwayWu/ImageForensicsOSN} \footnote{https://github.com/yhZhai/WSCL} \footnote{ https://github.com/mjkwon2021/CAT-Net} \footnote{ https://github.com/grip-unina/TruFor}. To ensure the fair comparison, we also \textit{retrain} PSCC-Net, MVSS-Net, IF-OSN, and WSCL on the training dataset of CAT-Net, in addition to directly using their released versions. We also involve two well-known clustering-based algorithms Lyu-NOI~\cite{lyu2014exposing} and PCA-NOI~\cite{trad_clustering2017} \footnote{NOI2\&5 in https://github.com/MKLab-ITI/image-forensics} as competitors..

\subsubsection{Evaluation Metrics} Follow the convention \cite{mvssnet, catnet, ifosn, trufor, wscl}, we utilize the pixel-level F1 and Intersection over Union (IoU) scores as the fixed-threshold metrics (higher the better), where the threshold is set to 0.5 by default. Formally, the macro-averaged F1 is defined as
\begin{equation}
	\mathrm{F1} = \frac{1}{Y}\sum_{y=1}^Y\frac{2 \times \mathrm{TP}_y}{2 \times \mathrm{TP}_y + \mathrm{FP}_y + \mathrm{FN}_y},
\end{equation}
where $\mathrm{TP}_y$, $\mathrm{FP}_y$, and $\mathrm{FN}_y$ represent True Positive, False Positive, and False Negative for a given class $y$ (``pristine'' or ``forged''), respectively. 

The IoU can be calculated as follows:
\begin{equation}
	\mathrm{IoU} = \frac{\mathbf{P} \cap \mathbf{Y}}{\mathbf{P} \cup \mathbf{Y}},
\end{equation}
where $\mathbf{P}$ and $\mathbf{Y}$ are the prediction and ground-truth masks, respectively. As for the threshold-agnostic metric, we adopt Area Under the Curve (AUC) following convention. Please refer to our code for the specific implementation, which is mainly based on the scikit-learn \cite{scikit-learn} extension package.

\subsubsection{Implementation Details} We implement FOCAL by using PyTorch deep learning framework. HRNet \cite{hrnet} and ViT \cite{vit} are adopted for the specific backbones of the FOCAL extractor. The Adam \cite{kingma2014adam} with default parameters is selected as the optimizer, and the learning rate is initialized to 1e-4. The batch size is set to 4 and the training is performed on 4 NVIDIA A100 GPU 40GB. All the input images are resized to $1024 \times 1024$, and the corresponding feature space of $\mathbf{F}$ is $\mathbb{R}^{256 \times 256 \times 256}$ for HRNet and $\mathbb{R}^{128 \times 128 \times 512}$ for ViT. Consistent with the existing competitors, data augmentation including random compression, scaling, blurring, and additive noise are applied during training to simulate real-world variations and enhance robustness.

\begin{figure*}[t!]
	\begin{center}
		\includegraphics[width=0.99\linewidth]{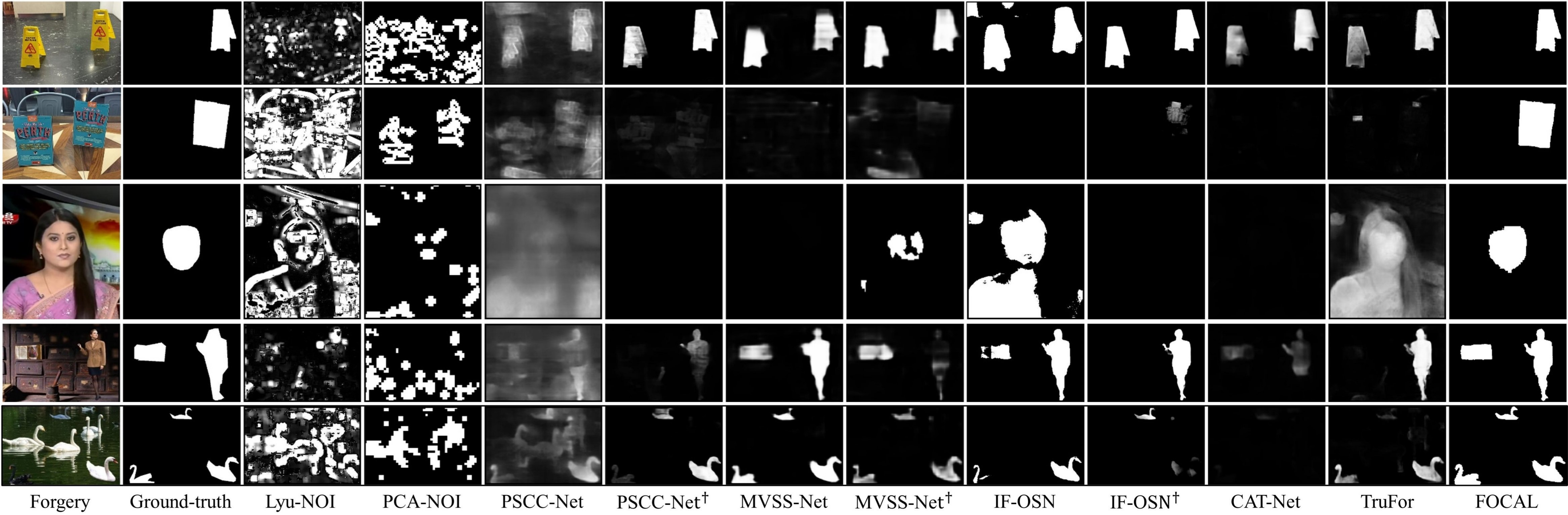}
	\end{center}
	\caption{Qualitative comparison of forgery detection results on some representative testing images. For each row, the images from left to right are forgery (input), ground-truth forgery mask, detection results generated by Lyu-NOI, PCA-NOI, PSCC-Net, MVSS-Net, IF-OSN, CAT-Net, TruFor and our FOCAL (Fusion), respectively. $^\dagger$: retrained versions with CAT-Net datasets.}
	\label{fig:cmp}
\end{figure*}

\subsection{Quantitative Comparisons}
Table~\ref{tab:cmp} lists quantitative comparisons of different image forgery detection methods, in terms of pixel-level F1 and IoU scores. Here we additionally report the results of PSCC-Net, MVSS-Net, IF-OSN, and WSCL retrained with the training set of CAT-Net. Generally, the retrained MVSS-Net and IF-OSN achieve comparable performance to their officially released versions, while the retrained PSCC-Net and WSCL lead to much better performance, \ie, +9.5\% in F1 and +5.0\% in IoU for PSCC-Net. This phenomenon indicates that different training datasets may have a huge impact on the eventual performance. For the benefits of the competing methods, we take the higher performance of the original and retrained versions in the following analysis.

\begin{table}[t]
	\caption{Quantitative comparison using pixel-level AUC as criterion.}
	\centering
	\scalebox{0.75}{
		\begin{tabular}{l|cccccc|c}
			\hline
			\hline
			Methods & \texttt{Columbia} & \texttt{Coverage} & \texttt{CASIA} & \texttt{MISD} & \texttt{NIST} & \texttt{FF++} & Mean\\
			\hline
			PSCC-Net \cite{psccnet} & .924 & .826 & .810 & .788 & .707 & .549 & .767 \\
			MVSS-Net \cite{mvssnet} & .881 & .803 & .835 & .795 & .725 & .662 & .784 \\
			TruFor \cite{trufor} & .860 & .871 & .904 & .832 & .796 & .857 & .853 \\
			\hline
			FOCAL & \textbf{.983} & \textbf{.920} & \textbf{.931} & \textbf{.890} & \textbf{.825} & \textbf{.937} & \textbf{.914} \\
			\hline
			\hline
		\end{tabular}
	}
	\label{tab:pixel-auc}
\end{table}

As can be observed from Table~\ref{tab:cmp}, the traditional clustering-based algorithms Lyu-NOI and PCA-NOI achieve unsatisfactory performance of $\sim$50\% in F1 and $\sim$11\% in IoU. This is mainly due to the fact that their hand-crafted noise features are heavily corrupted by post-processing operations commonly observed in the testing datasets. In contrast, the latest classification-based competitors with deep learning offer much better detection results. Among them, MVSS-Net and IF-OSN achieve slightly better results on \texttt{Columbia} and \texttt{MISD} datasets, with IoU scores being 78.4\% and 54.8\%, respectively; while on \texttt{CASIA} dataset, CAT-Net exhibits the better performance of IoU scores 64.2\%. The recently published TruFor achieves good results on the remaining three datasets. Thanks to the designed SCL in the training paradigm and the unsupervised clustering used in the testing, our FOCAL, whether utilizing single extractor (HRNet or ViT) or fused (HRNet + ViT), consistently leads to the best performance over all testing datasets in both F1 and IoU criteria. Particularly, FOCAL (Fusion) is shown to be effective in further boosting the performance (\eg, +10.6\% and +4.8\% IoU on \texttt{FF++} and \texttt{Coverage} respectively), via simple feature-level concatenation without the need of retraining. As can also be observed, FOCAL (Fusion) can avoid the bias of a single extractor backbone on some testing examples. Overall, FOCAL (Fusion) surpasses the best competing algorithm by big margins, \eg, +24.8\%, +18.9\%, +17.3\%, +15.3\%, +15.0\%, and +10.5\% in IoU on datasets \texttt{Coverage}, \texttt{Columbia}, \texttt{FF++}, \texttt{MISD}, \texttt{CASIA}, and \texttt{NIST}, respectively.

Table~\ref{tab:pixel-auc} additionally lists quantitative comparisons in terms of pixel-level AUC, to comprehensively evaluate threshold-agnostic separability between pristine and forged pixels. The reported AUC score of 91.4\% for FOCAL demonstrates superior discriminative capability compared to prior methods~\cite{trufor, mvssnet, psccnet} (76.7\%$\sim$85.3\%), validating that our method maintains robust forensic distinction across all decision thresholds rather than excelling only at a specific operating point.

Considering that the above evaluations are all based on pixel-level metrics, we calculate image-level forgery detection scores and report in Table~\ref{tab:image-level} in terms of AUC and accuracy metrics. Specifically, this calculation is implemented by deriving image-level forgery detection scores through the homogeneity analysis of our forensic-aware features $\mathbf{F}$, where normalized mutual information between local feature distributions serves as an indicator of global authenticity. As can be seen, Table~\ref{tab:image-level} demonstrates the superiority of FOCAL, with an improvement of 5.3\% in AUC and 5.8\% in accuracy over the second-best method. This capability emerges naturally from our localization-focused framework since forged images inherently exhibit feature heterogeneity between manipulated and pristine regions, enabling joint optimization of both pixel- and image-level detection without architectural modification.

\begin{table}[t]
	\caption{Quantitative comparison using image-level criteria.}
	\centering
	\scalebox{0.72}{
		\begin{tabular}{l|cc|cc|cc|cc|cc}
			\hline
			\hline
			\multirow{2}{*}{Methods} & \multicolumn{2}{c|}{\texttt{Coverage}} & \multicolumn{2}{c|}{\texttt{CASIA}} & \multicolumn{2}{c|}{\texttt{MISD}} & \multicolumn{2}{c|}{\texttt{FF++}} & \multicolumn{2}{c}{Mean} \\
			& AUC & Acc & AUC & Acc & AUC & Acc & AUC & Acc & AUC & Acc \\
			\hline
			PSCC-Net \cite{psccnet} & .657 & .550 & .869 & .683 & .712 & .594 & .764 & .736 & .751 & .641 \\
			MVSS-Net \cite{mvssnet} & .733 & .545 & .932 & .808 & .735 & .585 & .794 & .723 & .799 & .665 \\
			TruFor \cite{mvssnet} & .770 & .680 & .916 & .813 & .742 & .624 & .848 & .766 & .819 & .721 \\
			\hline
			FOCAL & \textbf{.823} & \textbf{.739} & \textbf{.962} & \textbf{.871} & \textbf{.814} & \textbf{.687} & \textbf{.890} & \textbf{.818} & \textbf{.872} & \textbf{.779} \\
			\hline
			\hline
		\end{tabular}
	}
	\label{tab:image-level}
\end{table}

\subsection{Qualitative Comparisons}
Fig.~\ref{fig:cmp} presents forgery detection results on some representative testing images. As can be noticed, traditional clustering-based methods with hand-crafted noise features Lyu-NOI and PCA-NOI perform poorly; many forged regions cannot be detected and a large number of false alarms exist. The classification-based method PSCC-Net also does not perform satisfactorily on these cross-domain testing data, where most of the forged regions are not detected. Similarly, CAT-Net and MVSS-Net miss many forged regions, resulting in inaccurate detection. TruFor and IF-OSN are slightly better in some examples; but many forged regions cannot be accurately identified and many untouched regions are falsely detected as forged. In contrast, our FOCAL (Fusion) not only accurately detects forged regions but also performs rather stably on cross-domain testing. Also, the false alarms have been remarkably suppressed.

Recall FOCAL implicitly assumes that all forged regions within one single image share similar features, though they could be made with different types of forgery (\eg, splicing and inpainting). An interesting question arising is whether FOCAL can detect multiple types of forgery simultaneously. The answer is affirmative. The examples shown in the last two rows of Fig.~\ref{fig:cmp} are from the \texttt{MISD} dataset, where multi-source splicing forgery is used. It can be noticed that FOCAL can still produce satisfactory detection results. The reason for the success in this challenging and practical scenario may be that there are more pristine than forged regions. The cluster with the largest amount of data will be directly marked as pristine, while the clusters with the smaller amount of data will be merged and all marked as forged. 

\begin{figure*}[!th]
	\begin{center}
		\includegraphics[width=0.98\linewidth]{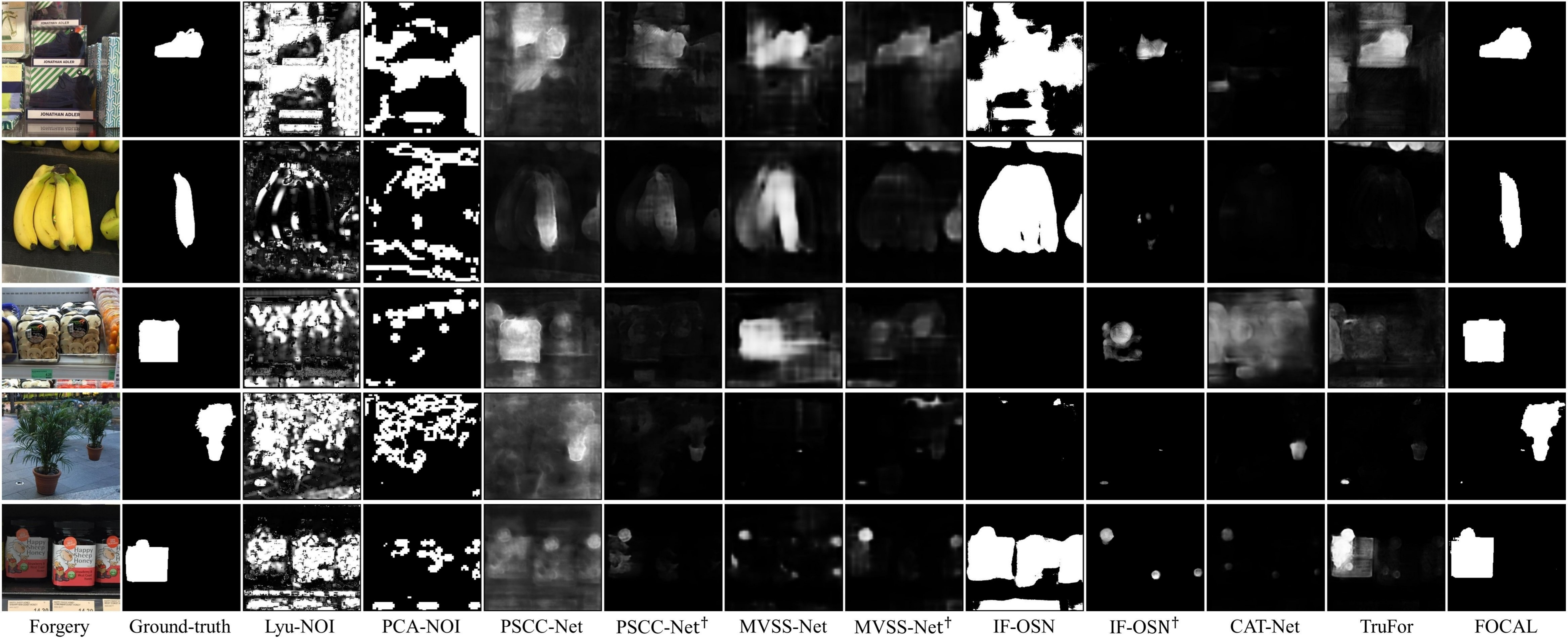}
	\end{center}
	\caption{Qualitative comparisons on \texttt{Coverage} \cite{coverage} dataset.}
	\vspace{1.2em}
	\label{fig:cmp_coverage}
\end{figure*}

\begin{figure*}[!th]
	\begin{center}
		\includegraphics[width=0.98\linewidth]{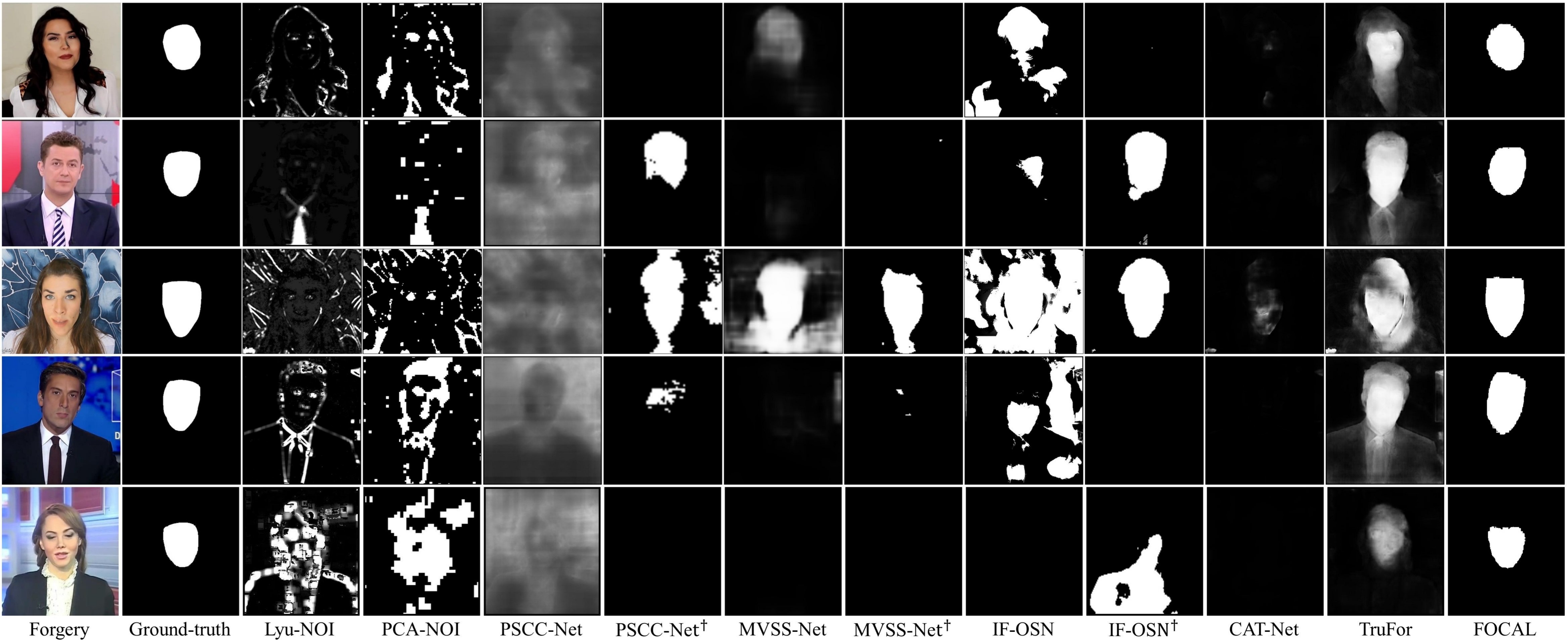}
	\end{center}
	\caption{Qualitative comparisons on \texttt{FF++} \cite{faceforensics2019} dataset.}
	\label{fig:cmp_ffpp}
\end{figure*}

More comparisons over testing datasets \texttt{Coverage} \cite{coverage}, \texttt{Columbia} \cite{columbia2006}, \texttt{NIST} \cite{nist2016}, \texttt{CASIA} \cite{casia2013}, \texttt{MISD} \cite{multisp}, \texttt{FF++} \cite{faceforensics2019} are given in Figs.~\ref{fig:cmp_coverage}$\sim$\ref{fig:cmp_ffpp}, respectively.

\begin{table}
	\centering
	\caption{Ablation studies regarding the backbone (Criterion is F1).}
	\scalebox{0.95}{
		\begin{tabular}{c|c|cccc|c}
			\hline
			\hline
			\multirow{2}{*}{Framework} & \multirow{2}{*}{Extractor} & \multicolumn{4}{c|}{Testing Datasets} & \multirow{2}{*}{Mean} \\
			\cline{3-6} & & \texttt{CASIA} & \texttt{MISD} & \texttt{NIST} & \texttt{FF++} &\\
			\cline{1-7} Traditional & HRNet & .718 & .727 & .569 & .675 & .672 \\
			\hline
			\multirow{4}{*}{\shortstack{FOCAL\\(Single)}} & HRNet & .866 & .851 & .708 & .835 & .815   \\
			& ViT & .889 & .875 & .725 & .848 & .834  \\
			& MiT & .684 & .732 & .653 & .787 & .714  \\
			& ConvNeXt & .465 & .538 & .462 & .575 & .510  \\
			\hline
			\multirow{2}{*}{\shortstack{FOCAL\\(Fusion)}} & \multirow{2}{*}{HRNet+ViT} & \multirow{2}{*}{\textbf{.904}} & \multirow{2}{*}{\textbf{.892}} & \multirow{2}{*}{\textbf{.741}} & \multirow{2}{*}{\textbf{.902}} & \multirow{2}{*}{\textbf{.860}}\\
			& & & & & & \\
			\hline
			\hline
		\end{tabular}
	}
	\label{tab:ablation_extractor}
\end{table}

\subsection{Ablation Studies}

We now analyze how each component contributes to the FOCAL framework in terms of extractor backbone, loss function, and clustering algorithm.

\begin{figure}[t!]
	\begin{center}
		\includegraphics[width=0.96\linewidth]{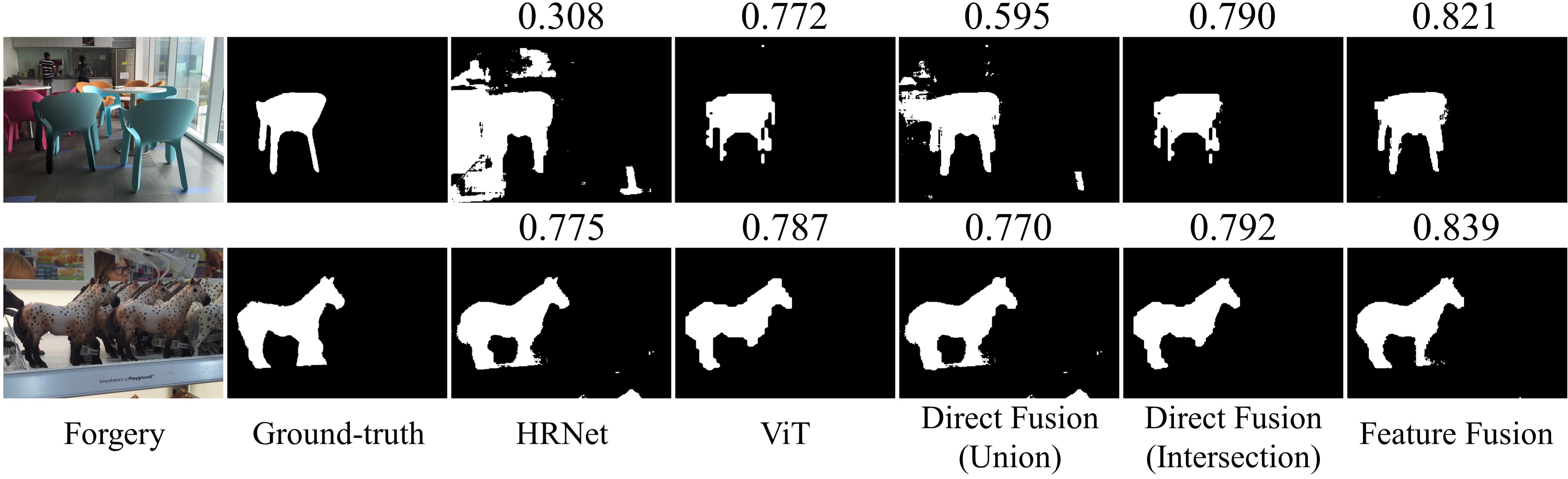}
	\end{center}
	\caption{Impact of different extractors and fusion strategies. The number above the mask represents the score of IoU.}
	\label{fig:fusion}
\end{figure}

\subsubsection{Extractor Backbones}
We start ablation studies with the selection of the FOCAL extractor, where the key point is how to select the backbone. Since our focus is \textit{not} to design a brand new backbone, we directly adopt the most commonly-used backbones proposed in recent years, namely, HRNet \cite{hrnet}, ConvNeXt \cite{convnext}, ViT \cite{vit}, and MiT \cite{mit}, for the comparison. The corresponding detection results are shown in Table~\ref{tab:ablation_extractor}, where the first row gives the performance of traditional classification-based framework as a comparison. As can be observed, ViT leads to the superior detection performance among these compared backbones. This might be due to its attention ability to extract richer forgery features by globally modelling long-range dependencies. Also, note that the extractor backbone in FOCAL can be flexibly replaced by a more advanced architecture when it is available. Further, in Fig.~\ref{fig:fusion}, we compare the visual results of the predicted forgery masks when using different backbones and fusion strategies. It can be seen that the feature fusion surpasses naive result-level fusions (\ie, union or intersection fusions) by a significant margin.    

\begin{table}
	\centering
	\caption{Ablation studies regarding the loss function (Criterion is F1).}
	\scalebox{0.78}{
		\begin{tabular}{lcc||cccc|c}
			\hline
			\hline
			\multirow{2}{*}{Loss} & \multirow{2}{*}{$w_{ij}$} & Image & \multicolumn{4}{c|}{Testing Datasets} & \multirow{2}{*}{Mean} \\
			\cline{4-7} & & Manner & \texttt{CASIA} & \texttt{MISD} & \texttt{NIST} & \texttt{FF++} & \\
			\cline{1-8} Triplet & - & - & .408 & .497 & .411 & .459 & .444  \\
			DCL & - & - & .514 & .618 & .581 & .547 & .565  \\
			Circle & - & - & .763 & .745 & .637 & .725 & .718  \\
			NCE & - & - & .787 & .753 & .642 & .743 & .731  \\
			\hline
			$\mathcal{L}_{\mathrm{SCL}}$ & - & - & .799 & .764 & .633 & .752 & .737  \\
			$\mathcal{L}_{\mathrm{SCL}}$ & \checkmark & - & .854 & .831 & .674 & .861 & .805  \\
			$\mathcal{L}_{\mathrm{SCL}}$ & \checkmark & \checkmark & \textbf{.904} & \textbf{.892} & \textbf{.741} & \textbf{.902} & \textbf{.860}\\
			\hline
			\hline
		\end{tabular}
	}
	\label{tab:ablation_loss}
\end{table}

\subsubsection{Loss Functions}

The soft contrastive learning module plays a crucial role in FOCAL. We now evaluate the performance of FOCAL variants by replacing our adopted $\mathcal{L}_{\mathrm{SCL}}$ with existing contrastive losses, such as Triplet \cite{triplet}, DCL \cite{yeh2022decoupled}, Circle \cite{circleloss}, and the original NCE \cite{infonce}. As can be seen from Table~\ref{tab:ablation_loss}, not all of these loss functions are suitable for the forgery detection task. For example, Triplet restricts an equal penalty strength to the distance score of every query positive or negative pair \cite{circleloss}, which results in the model collapse. By re-weighting each distance score under supervision, Circle has a more flexible optimization and definite convergence target, far exceeding Triplet. Although DCL and NCE have the same supervision mechanism, the positive constraint removed by DCL makes it easy to get stuck in poor local optima, causing the performance of DCL far inferior to that of NCE. Additionally, by introducing optimizable coefficients $w_{ij}$ to weight features, our proposed $\mathcal{L}_{\mathrm{SCL}}$ can better cope with label ambiguity, resulting in a 7.4\% F1 gains compared with vanilla NCE. Note that when the weighting provided by $w_{ij}$ is dropped, the supervision of $\mathcal{L}_{\mathrm{SCL}}$ degenerates to that of NCE, which emphasizes the importance of our proposed SCL strategy. Finally, we give the results when the overall $\mathcal{L}_{\mathrm{SCL}}$ is computed in an image-by-image manner rather than batch-level (last two rows in Table~\ref{tab:ablation_loss}). As expected, our image-by-image overall loss design significantly outperforms such a batch-level loss (+5.5\% in F1). The big performance gap further indicates the necessity of explicitly using the relative definition of forged and pristine pixels within one single image.

\begin{figure*}[t!]
	\begin{center}
		\includegraphics[width=0.98\linewidth]{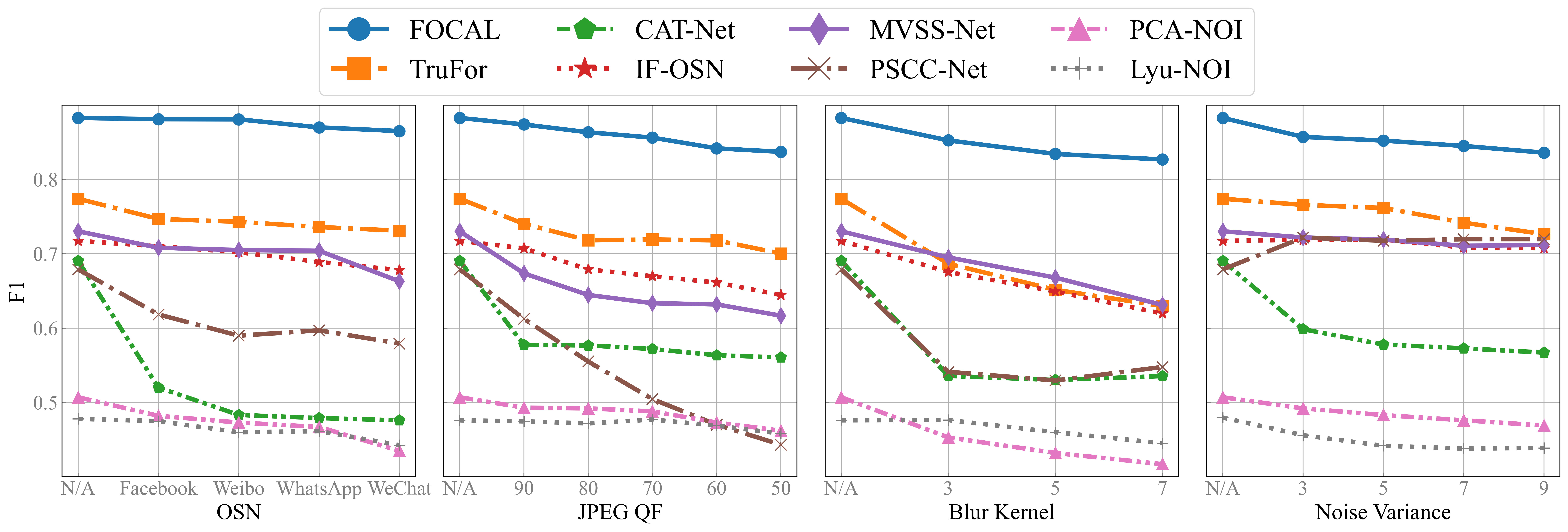}
	\end{center}
	\caption{Robustness evaluations against OSN transmission, JPEG compression, Gaussian blurring and Gaussian noise addition.}
	\label{fig:post_cmp}
\end{figure*}

\subsubsection{Clustering Algorithms}\label{sec:clustering}

Apart from contrastive learning, another key module of FOCAL is the clustering algorithm for generating the final predicted forgery mask. To explore the most suitable clustering algorithm for the FOCAL framework, we evaluate the most popular clustering algorithms, K-means~\cite{kmeans}, B-K-means~\cite{bilge2013scalable}, BIRCH~\cite{zhang1996birch}, Hierarchical~\cite{ward1963hierarchical} and HDBSCAN~\cite{ester1996density}, and report the results in Table~\ref{tab:ablation_cluster}. 

\begin{figure}[t!]
	\begin{center}
		\includegraphics[width=0.98\linewidth]{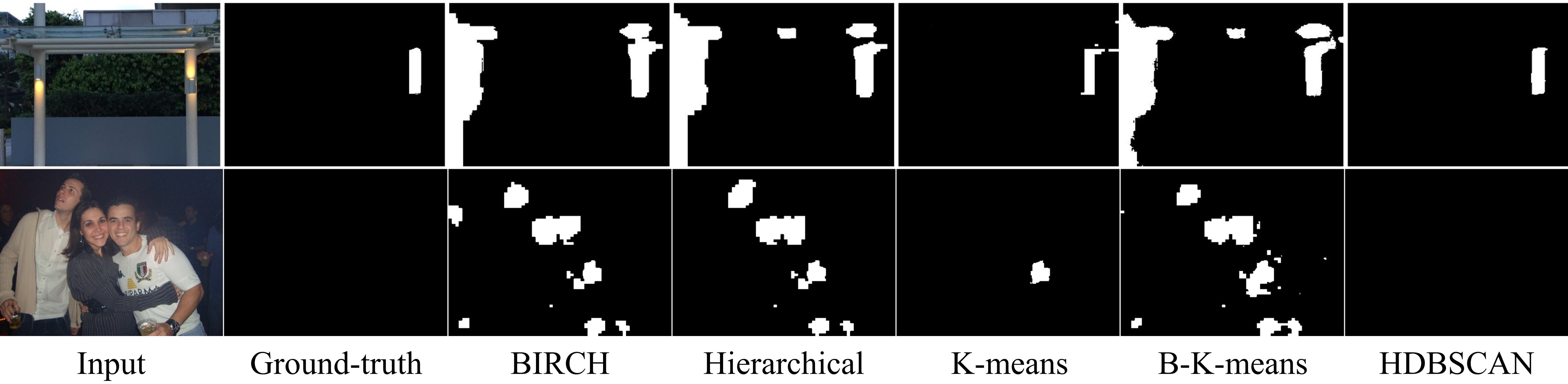}
	\end{center}
	\caption{Impact of different clustering algorithms.}
	\label{fig:cluster_vis_cmp}
\end{figure}

For K-means, B-K-means, and BIRCH algorithms, the number of clusters to be formed is set to 2, while other parameters take their default values. As can be observed, the aforementioned algorithms exhibit comparable performance, attributed to the discriminative nature of the features $\mathbf{F}$ learned by the extractor. Among them, HDBSCAN performs the best, surpassing the second-place one by 1.4\% F1. Recalling that $\mathbf{F}$ could have $256\times256=65536$ elements to be clustered. Those clustering algorithms such as spectral clustering \cite{shi2000normalized} and affinity propagation \cite{frey2007clustering} that cannot be extended to large-scale elements are extremely slow and thereby omitted. We also would like to point out a potential limitation of using fixed numbers of clusters in K-means, B-K-means, and BIRCH. For completely pristine images (no forged regions), these clustering methods still force to produce two clusters, inevitably resulting in false alarms (see the last row in Fig.~\ref{fig:cluster_vis_cmp}). 

To further evaluate false alarms (lower the better) on pristine images, we conduct additional experiments on several pristine dataset, \texttt{ImageNet}~\cite{deng2009imagenet}, \texttt{COCO}~\cite{coco2014}, and \texttt{VISION}~\cite{vision2017} by randomly sampling 2000 images each. As shown in Table~\ref{tab:false_alarm}, the density-based algorithm HDBSCAN adopted by FOCAL can dynamically determine the number of final clusters, effectively suppressing the false alarms by 4.1\% for pristine images. Furthermore, our FOCAL (even using K-means) still outperforms the competitors \cite{mvssnet, catnet, trufor} by big margins.

\begin{table}[t]
	\centering
	\caption{Ablation studies regarding the clustering (Criterion is F1).}
	\scalebox{0.9}{
		\begin{tabular}{l|cccc|c}
			\hline
			\hline
			\multirow{2}{*}{Clustering} & \multicolumn{4}{c|}{Testing Datasets} & \multirow{2}{*}{Mean} \\
			\cline{2-5} & \texttt{CASIA} & \texttt{MISD} & \texttt{NIST} & \texttt{FF++} & \\
			\cline{1-6} BIRCH \cite{zhang1996birch} & .876 & .854 & .712 & .868 & .828  \\
			Hierarchical \cite{ward1963hierarchical} & .880 & .859 & .720 & .870 & .832  \\
			K-means \cite{kmeans} & .891 & .876 & .727 & .889 & .846  \\
			B-K-means \cite{bilge2013scalable} & .866 & .873 & .725 & .892 & .839  \\
			HDBSCAN \cite{ester1996density} & \textbf{.904} & \textbf{.892} & \textbf{.741} & \textbf{.902} & \textbf{.860} \\
			\hline
			\hline
		\end{tabular}
	}
	\label{tab:ablation_cluster}
\end{table}

\subsection{Robustness Evaluation}

The forged images often undergo a series of post-processing operations, such as compression, blurring, noise addition, attempting to eliminate forgery traces or mislead forgery detection algorithms. In addition, online social networks (OSNs), as prevailing transmission channels for images, have been shown to seriously affect image forensic algorithms \cite{ifosn}. It is therefore important to evaluate the robustness of all competing algorithms against post-processing operations and OSN transmission. Specifically, we apply the aforementioned degradation to the original testing datasets, and plot the results in Fig.~\ref{fig:post_cmp}. It can be observed that although CAT-Net~\cite{catnet} achieves good performance on the original dataset, it is vulnerable to post-processing and OSN transmissions. TruFor~\cite{trufor}, MVSS-Net~\cite{mvssnet} and IF-OSN~\cite{ifosn} exhibit certain degrees of robustness against these distortions. In contrast, our FOCAL still consistently achieves the best performance and robustness over these competitors. For instance, FOCAL only suffers $\sim$0.2\% performance degradation against Facebook or Weibo transmissions.

\begin{table}[t]
	\centering
	\caption{False alarm rate (\%) on pristine datasets.}
	\scalebox{0.9}{
		\begin{tabular}{lc|ccc|c}
			\hline
			\hline
			\multirow{2}{*}{Methods} & \multirow{2}{*}{Clustering}&  \multicolumn{3}{c|}{Pristine Datasets} & \multirow{2}{*}{Mean} \\
			\cline{3-5} & & \texttt{ImageNet} & \texttt{COCO} & \texttt{VISION} & \\
			\cline{1-6} MVSS-Net \cite{mvssnet} & - & .229 & .185 & .132 & .182 \\
			CAT-Net \cite{catnet} & - & .208 & .154 & .179 & .180 \\
			TruFor \cite{trufor} & - & .143 & .131 & .124 & .133 \\
			\hline
			FOCAL & K-means & .125 & .117 & .106 & .116 \\
			FOCAL & HDBSCAN & \textbf{.086} & \textbf{.091} & \textbf{.049} & \textbf{.075} \\
			\hline
			\hline
		\end{tabular}
	}
	\label{tab:false_alarm}
\end{table}

\section{Conclusion}\label{sec:conclusion}

We have explicitly pointed out the importance of the relative definition of forged and pristine pixels within an image, which has been severely overlooked by existing forgery detection methods. Inspired by this rethinking, we have proposed FOCAL, a novel, simple yet effective image forgery detection framework, based on SCL supervision in an image-by-image manner and unsupervised clustering. Extensive experiments have been given to demonstrate our superior performance.

\begin{figure*}[!th]
	\begin{center}
		\includegraphics[width=0.98\linewidth]{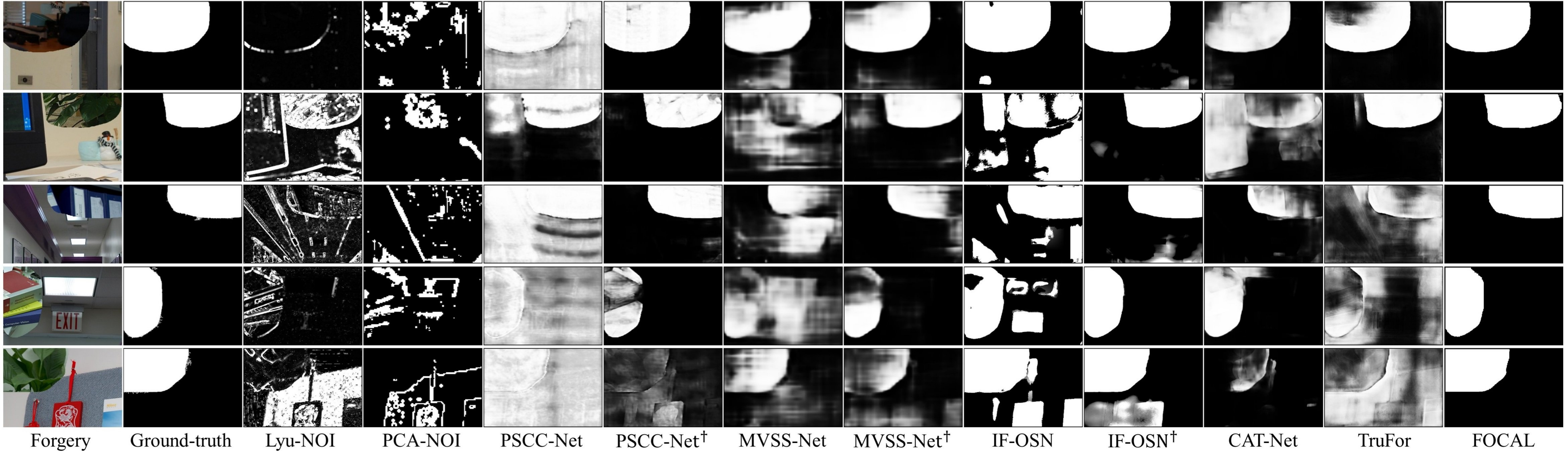}
	\end{center}
	\caption{Qualitative comparisons on \texttt{Columbia} \cite{columbia2006} dataset.}
	\vspace{1.2em}
	\label{fig:cmp_columbia}
\end{figure*}

\begin{figure*}[!th]
	\begin{center}
		\includegraphics[width=0.98\linewidth]{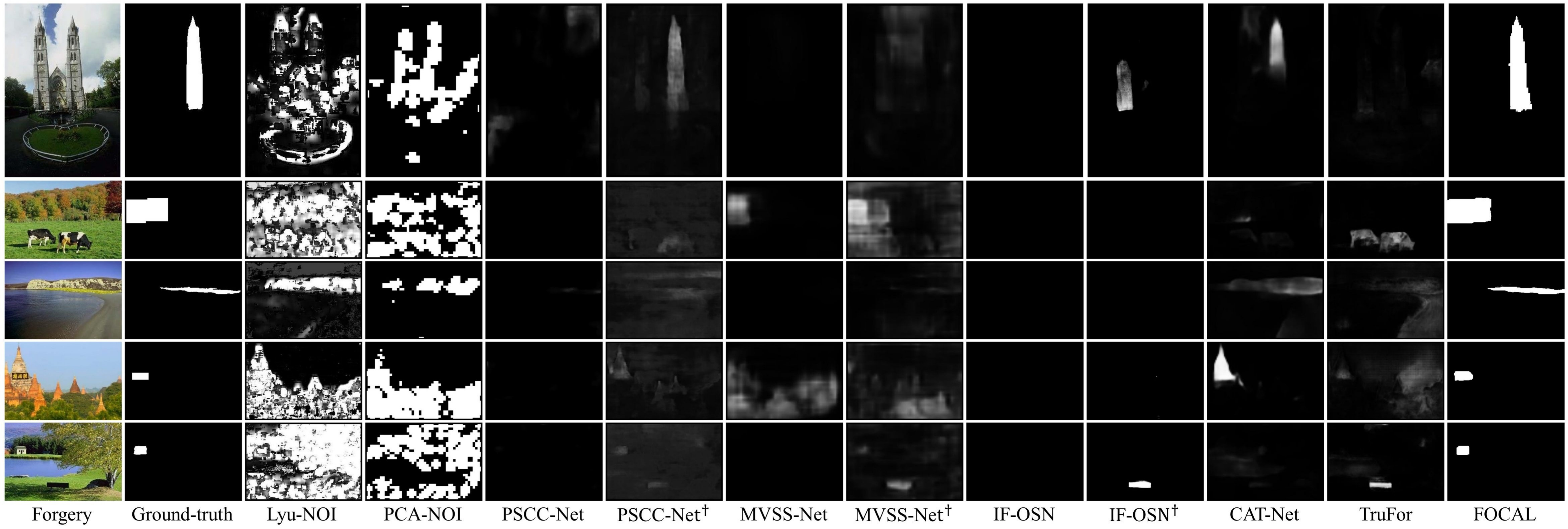}
	\end{center}
	\caption{Qualitative comparisons on \texttt{CASIA} \cite{casia2013} dataset.}
	\vspace{1.2em}
	\label{fig:cmp_casia}
\end{figure*}

\begin{figure*}[!th]
	\begin{center}
		\includegraphics[width=0.98\linewidth]{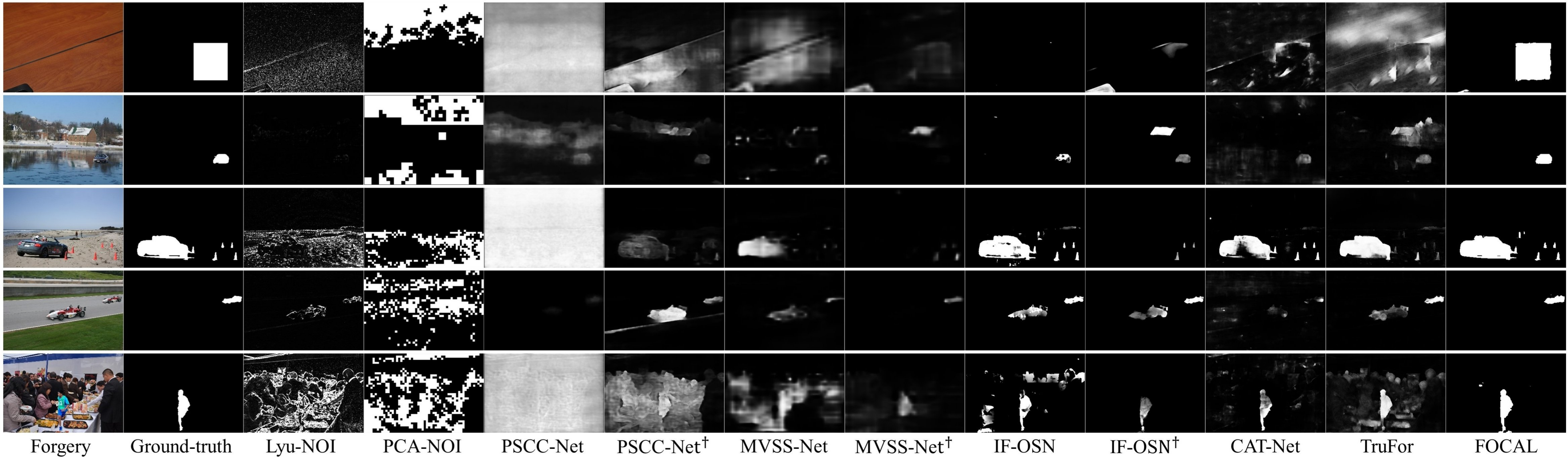}
	\end{center}
	\caption{Qualitative comparisons on \texttt{NIST} \cite{nist2016} dataset.}
	\label{fig:cmp_nist}
\end{figure*}

\begin{figure*}[!th]
	\begin{center}
		\includegraphics[width=0.98\linewidth]{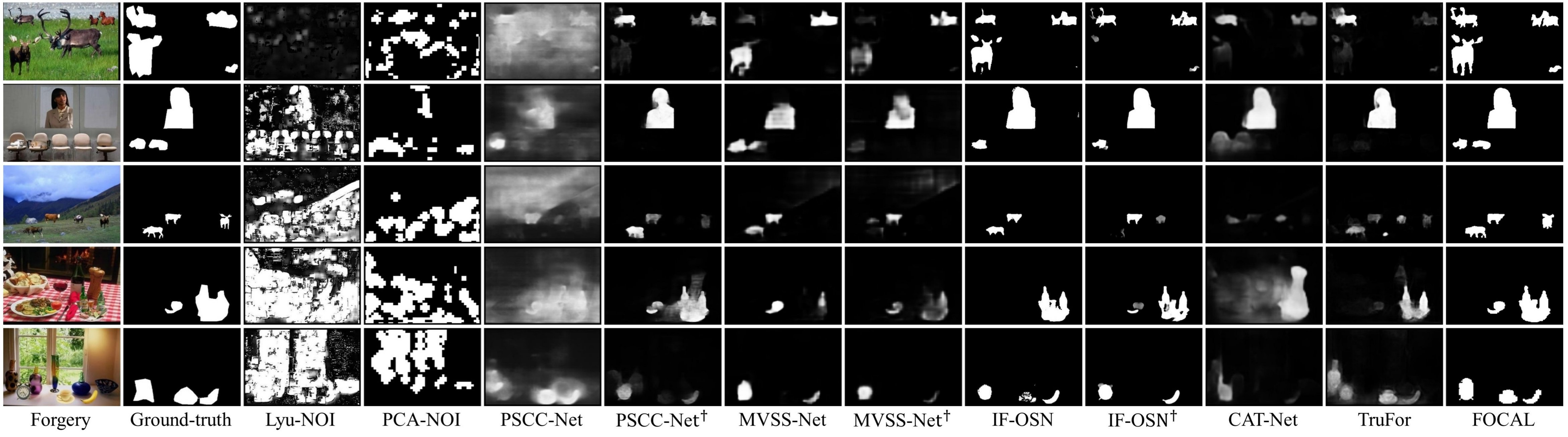}
	\end{center}
	\caption{Qualitative comparisons on \texttt{MISD} \cite{multisp} dataset.}
	\label{fig:cmp_multisp}
\end{figure*}

{
\bibliographystyle{IEEEtran}
\bibliography{ref}
}

\begin{IEEEbiography}[{\includegraphics[width=1in,height=1.25in,clip,keepaspectratio]{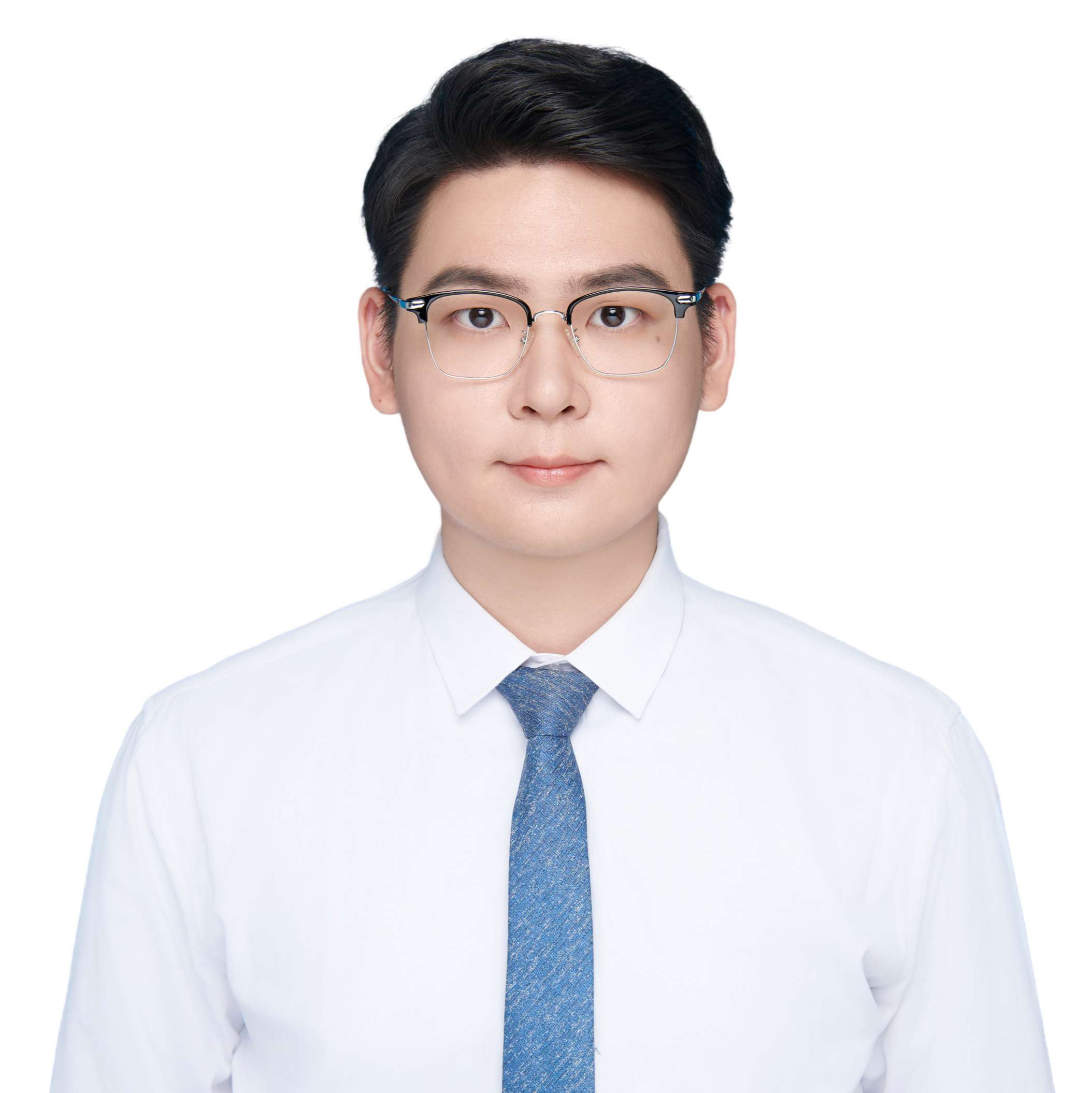}}]{Haiwei Wu} (Member, IEEE) received the B.S., M.S., and Ph.D. degrees in computer science from University of Macau, Macau, China, in 2018, 2020 and 2023, respectively. From 2023 to 2024, he was a postdoctoral research fellow at City University of Hong Kong. He is currently a Professor with the School of Computer Science and Engineering, University of Electronic Science and Technology of China. His research interests include multimedia security, image processing, and trustworthy AI.
\end{IEEEbiography}

\begin{IEEEbiography}[{\includegraphics[width=1in,height=1.25in,clip,keepaspectratio]{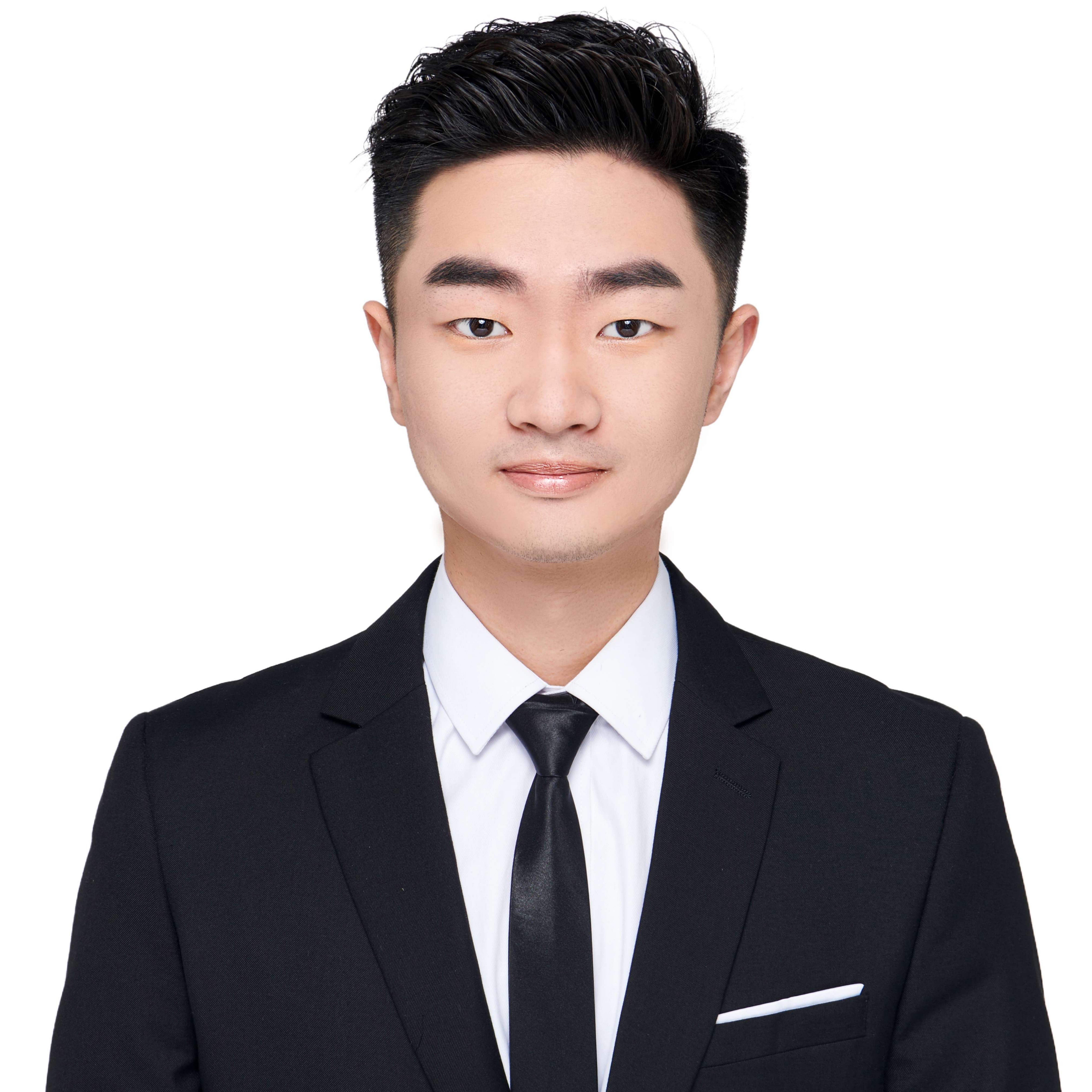}}]{Yiming Chen} (Student Member, IEEE) received the B.S. degree from the University of Electronic Science and Technology of China, in 2019, and the M.S. degree from the National University of Singapore, in 2020. He is currently pursuing the Ph.D. degree in University of Macau, China. His research interests include AI security and backdoor vulnerabilities of AI Models.
\end{IEEEbiography}

\begin{IEEEbiography}[{\includegraphics[width=1in,height=1.25in,clip,keepaspectratio]{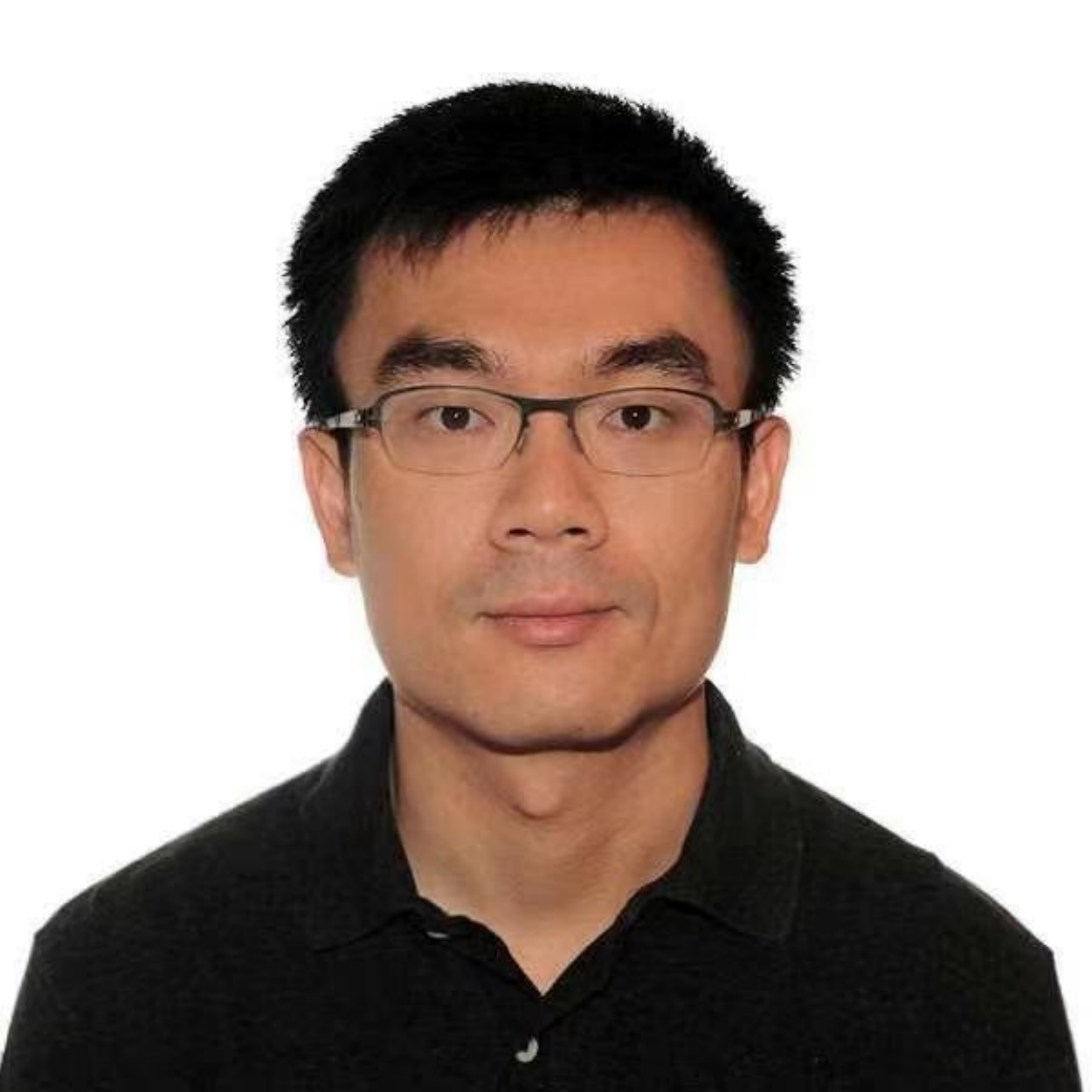}}]{Jiantao Zhou} (Senior Member, IEEE) received the B.Eng. degree from the Department of Electronic Engineering, Dalian University of Technology, in 2002, the M.Phil. degree from the Department of Radio Engineering, Southeast University, in 2005, and the Ph.D. degree from the Department of Electronic and Computer Engineering, Hong Kong University of Science and Technology, in 2009. He held various research positions with the University of Illinois at Urbana-Champaign, Hong Kong University of Science and Technology, and McMaster University. He is now the Head and Professor with the Department of Computer and Information Science, Faculty of Science and Technology, University of Macau. His research interests include multimedia security and forensics, multimedia signal processing, artificial intelligence and big data. He holds four granted U.S. patents and two granted Chinese patents. He has coauthored two papers that received the Best Paper Award at the IEEE Pacific-Rim Conference on Multimedia in 2007 and the Best Student Paper Award at the IEEE International Conference on Multimedia and Expo in 2016. He is serving as an Associate Editor for the IEEE TRANSACTIONS ON IMAGE PROCESSING, the IEEE TRANSACTIONS ON MULTIMEDIA, and the IEEE TRANSACTIONS ON DEPENDABLE and SECURE COMPUTING.
\end{IEEEbiography}

\begin{IEEEbiography}[{\includegraphics[width=1in,height=1.25in,clip,keepaspectratio]{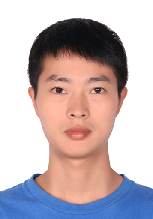}}]{Yuanman Li} (Senior Member, IEEE) received the B.Eng. degree in software engineering from Chongqing University, Chongqing, China, in 2012, and the Ph.D. degree in computer science from University of Macau, Macau, 2018. From 2018 to 2019, he was a Post-doctoral Fellow with the State Key Laboratory of Internet of Things for Smart City, University of Macau. He is currently an Assistant Professor with the College of Electronics and Information Engineering, Shenzhen University, Shenzhen, China. His current research interests include data representation, computer vision, and multimedia security.
\end{IEEEbiography}

\end{document}